\documentclass[journal]{IEEEtran}
\usepackage{graphicx}
\usepackage{pifont}
\usepackage{soul}
\usepackage{url}
\usepackage{hyperref}
\usepackage{verbatim}
\usepackage{subfigure}
\usepackage{calc}
\usepackage{amstext,amsmath,amsthm,amssymb}
\usepackage{mathrsfs}
\usepackage{stackengine}
\newcommand\IncG[2][]{\addstackgap{%
\raisebox{-.5\height}{\includegraphics[#1]{#2}}}}
\usepackage{array}

\usepackage{stfloats}
\usepackage{xcolor}
\usepackage{paralist}
\usepackage{multirow}
\usepackage{makecell}
\usepackage{bm}
\usepackage{booktabs}

\usepackage{algorithm}  
\usepackage{algpseudocode}

\usepackage{anyfontsize}
\usepackage{eqparbox}
\newdimen{\algindent}
\algnewcommand\LeftComment[2]{%
\hspace{#1\algindent}$\triangleright$ \eqparbox{COMMENT}{#2} \hfill %
}

\usepackage{xspace}

\makeatletter
\DeclareRobustCommand\onedot{\futurelet\@let@token\@onedot}
\def\@onedot{\ifx\@let@token.\else.\null\fi\xspace}

\def\eg{\emph{e.g}\onedot} 
\def\ie{\emph{i.e}\onedot}

\makeatother

\ifCLASSINFOpdf
\else
\fi
\begin{document}
%
\title{Lifelong-MonoDepth: Lifelong Learning for Multi-Domain Monocular Metric Depth Estimation}
%
%
%

\author{Junjie Hu,~\IEEEmembership{Member,~IEEE,}
Chenyou Fan, 
Liguang Zhou,
Qing Gao, 
Honghai Liu,~\IEEEmembership{Fellow,~IEEE} \\
and Tin Lun Lam,~\IEEEmembership{Senior Member,~IEEE}
\IEEEcompsocitemizethanks{\IEEEcompsocthanksitem J.Hu, L.Zhou, and Tin Lun Lam are with the Shenzhen Institute of Artificial Intelligence and Robotics for Society (AIRS), The Chinese University of Hong Kong, Shenzhen, China.
E-mail: hujunjie@cuhk.edu.cn.
\IEEEcompsocthanksitem C.Fan is with the School of Artificial Intelligence, South China Normal University, China. E-mail: fanchenyou@scnu.edu.cn.
\IEEEcompsocthanksitem Q.Gao is with the School of Electronics and Communication Engineering, Sun Yat-sen University, Shenzhen, China.
E-mail: gaoqing.ieee@gmail.com.
\IEEEcompsocthanksitem H.Liu is with the State Key Laboratory of
Robotics and systems, Harbin Institute of Technology (Shenzhen), China. E-mail: honghai.liu@hit.edu.cn.
\IEEEcompsocthanksitem L.Zhou and T.L.Lam are with the School of Science and Engineering, The Chinese University of Hong Kong, Shenzhen, China.
E-mail:  liguangzhou@link.cuhk.edu.cn, tllam@cuhk.edu.cn.
\IEEEcompsocthanksitem T.L.Lam is the corresponding author.}

\thanks{
This work was supported in part by the National Natural Science Foundation of China under Grants 62073274 and 62106156, in part by Shenzhen Science and Technology Program under Grant JCYJ20220818103000001, and in part by the Shenzhen Institute of Artificial Intelligence and Robotics for Society under Grant AC01202101103.}
}

\maketitle


\begin{abstract}
With the rapid advancements in autonomous driving and robot navigation, there is a growing demand for lifelong learning models capable of estimating metric (absolute) depth. Lifelong learning approaches potentially offer significant cost savings in terms of model training, data storage, and collection. However, the quality of RGB images and depth maps is sensor-dependent, and depth maps in the real world exhibit domain-specific characteristics, leading to variations in depth ranges. These challenges limit existing methods to lifelong learning scenarios with small domain gaps and relative depth map estimation.
To facilitate lifelong metric depth learning, we identify three crucial technical challenges that require attention: i) developing a model capable of addressing the depth scale variation through scale-aware depth learning, ii) devising an effective learning strategy to handle significant domain gaps, and iii) creating an automated solution for domain-aware depth inference in practical applications.
Based on the aforementioned considerations, in this paper, we present i) a lightweight multi-head framework that effectively tackles the depth scale imbalance, ii) an uncertainty-aware lifelong learning solution that adeptly handles significant domain gaps, and iii) an online domain-specific predictor selection method for real-time inference. 
Through extensive numerical studies, we show that the proposed method can achieve good efficiency, stability, and plasticity, leading the benchmarks by 8\% $\sim$ 15\%. 
The code is available at \url{https://github.com/FreeformRobotics/Lifelong-MonoDepth}.

\end{abstract}


\section{Introduction}
\begin{figure}[t!]
    \centering
    \includegraphics[width = \linewidth]{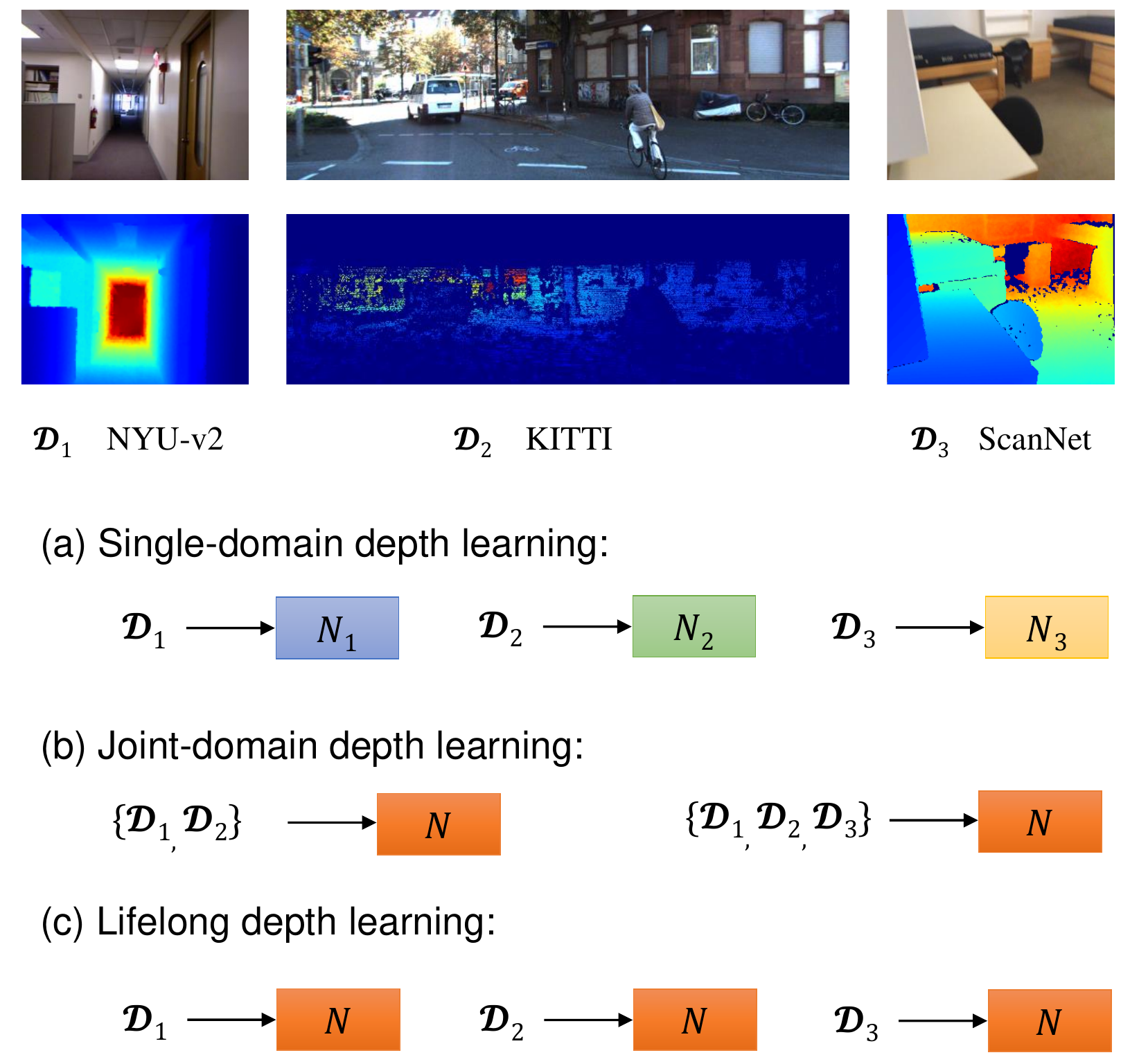}
    \vspace{-5mm}
    \caption{Depth learning in the real world where the same color and different colors denote the same and different models. Traditional approaches include single-domain learning to train a domain-specific model as (a), and joint-domain learning to obtain a domain-robust model as (b).
    We aim to learn a model that can infer metric depth maps for multiple domains in a lifelong learning manner as (c).}
    \label{fig-example}
\end{figure}

\IEEEPARstart{A}CQUIRING scene depths in real depth scale is an essential requirement for real-world applications, e.g., SLAM \cite{Tateno2017CNNSLAMRD}, self-driving \cite{song2021self}, robot navigation \cite{Mendes2021OnDL}, 3D reconstruction \cite{guizilini20203d}, human-computer interaction \cite{duan2012depth}, augmented reality \cite{Du2020DepthLab}, etc. As a cost-effective solution to depth sensors, monocular depth estimation (MDE) aims to infer depth maps from visual images. MDE has gained great success by learning with deep convolutional neural networks (CNNs) in a data-driven fashion.
In the early stage, traditional studies handled MDE by training and testing on a single-domain \cite{laina2016deeper,fu2018deep,ma2017sparse,Chen2019structure-aware,Hu2019RevisitingSI,Yin2019enforcing}, as shown in Fig.~\ref{fig-example}. (a).

However, learning-based methods have often been criticized and questioned due to their poor generalizability for out-of-distribution data. 
Despite the recent trend of tackling poor generalizability by covering possible domains as much as possible \cite{ranftl2020towards,ranftl2021vision,xian2018monocular,tro-depth}, as seen in Fig.~\ref{fig-example}. (b), it is impossible to exhaust all possible patterns of data in the real world.
When there are some new patterns of data or target domains, a pre-trained model has to be re-trained from scratch, resulting in a tremendous waste of time and cost.
Inspired by human cognition, researchers have attempted to empower CNNs with lifelong learning mechanisms which aim to perform incremental learning on new domains or tasks with the minimum increase over model complexity, training time, and reuse of data on old tasks.
This practice has already seen promising results on image recognition \cite{de2021continual,rusu2016progressive,kirkpatrick2017overcoming,aljundi2017expert}.

On the other hand, since there is a significant difference between image recognition and MDE, it is largely unknown how to enable lifelong learning for MDE (Fig.~\ref{fig-example}. (c)). Most previous approaches \cite{ranftl2020towards,ranftl2021vision,tro-depth} of multi-domain learning choose only to infer relative depth maps to tackle domain gaps. 
Besides, only a few studies \cite{kuznietsov2021comoda,Khan2021TowardsCO,zhang2020online} have tried empowering MDE with lifelong learning, and none of them can infer scale-aware metric depth maps.
In this paper, we extensively study this under-explored problem and provide some valuable insights. 
We identify two major challenges of scale-ware MDE that cause catastrophic forgetting (forgetting learned knowledge after updating a trained model on a new domain) when performing lifelong learning, including
\begin{itemize}
    \item Significant domain gap: both visual images and depth images are significantly different across different domains. Thus, a trained model on old domains tends to shift its parameters
to accommodate a new domain significantly.
    \item Depth scale imbalance: scene depth scales are usually domain-dependent and dominated by a specific range such that lifelong learning across two domains of different scales is ineffective.
\end{itemize}

The above challenges limit existing methods to lifelong learning scenarios with small domain gaps and relative depth map estimation, as demonstrated in \cite{zhang2020online,kuznietsov2021comoda,Khan2021TowardsCO}. To facilitate lifelong metric depth learning, we propose a general framework, {\it Lifelong-MonoDepth} for lifelong learning on MDE.  We posit that three pivotal technical challenges necessitate attention: i) developing a model capable of addressing the depth scale imbalance issue through scale-aware depth learning, ii) devising an effective learning strategy to handle significant domain gaps, and iii) creating an automated solution for domain-aware depth inference in practice. 

We consider MDE under natural circumstances where agents work in complex real-world environments, including both indoor and outdoor scenarios. 
Fig.~\ref{fig-example} shows several samples of images and depth maps from three different domains.
As seen, depth maps captured in the real world are significantly different across domains; their quality and scales are domain-dependent. 
Therefore, the model has to efficiently assemble multiple prediction branches for multi-domain metric depth inference.  Otherwise, if the model only has a single branch for outputting depth, its parameters will gradually shift to accommodate the depth range of a new domain when performing lifelong learning. Additionally, we need to also ensure the memory-efficient and computationally frugal computation for MDE task considering their real-world deployment.
To this end, we present a lightweight multi-head framework that consists of a domain-shared encoder and domain-specific layers. The framework dynamically grows a domain-specific predictor with only a 0.21M increase in parameters when learning on a new domain.
 The framework allows robust metric depth learning across multi-domains.

To facilitate effective and practical learning of our proposed model, we adhere to the principles of lifelong learning that minimizes the use of historical data
to reduce data storage costs and keeps the increase in parameters for new domains as minimal as possible to improve efficiency in practical applications.
To achieve this, we adopt a regularization term that applies a knowledge distillation loss as \cite{li2017learning}, thereby eliminating the need for extensive storage of historical data  as prior works \cite{kuznietsov2021comoda,Khan2021TowardsCO} and avoiding excessive parameter increases as \cite{zhang2020online}. To handle the significant domain gaps, we introduce an uncertainty-aware knowledge preservation solution by incorporating uncertainty estimation and uncertainty consistency into our lifelong learning framework. The former addresses potential performance degradation resulting from outliers present in ground-truth depth maps of certain domains, thereby achieving a better balance between different domains. The latter poses a strong regularization for preserving the knowledge of original domains.
 To further mitigate the significant domain gap, we also apply a replay loss using only 500 randomly selected image and depth pairs from each original domain. By strategically selecting this limited subset of data, we effectively bridge the domain gap while maintaining computational efficiency and ensuring reliable learning outcomes.

We then consider how to dynamically select the optimal domain-specific predictor given an input image during inference. We assume the input image belongs to one of the target domains, and the key is how to identify that domain. As the replay data is a small subset of each domain, we propose to compare the distance between the image and each domain in the feature space. Then, the closest domain is the one with the minimum distance.  

To validate the effectiveness of the proposed method, we perform lifelong depth learning on three real-world datasets with significant domain gaps. We show through experiments that the proposed method can i) enable lifelong learning for scale-aware depth estimation, ii) cope with significant domain shift during lifelong learning, and iii) infer a depth map in real-time.

In summary, our contributions are:
\begin{itemize}
    \item We present an efficient multi-head framework that enables lifelong, cross-domain, and scale-aware monocular depth learning. To our best knowledge, we are the first to fulfill multi-domain metric depth estimation via lifelong learning.
    \item We introduce an uncertainty-aware knowledge preservation solution by incorporating uncertainty estimation and uncertainty consistency into our lifelong learning framework. The former strikes a balance between different domains since the quality of depth is domain-dependent, and the latter provides a strong regularization for better preserving the model's knowledge on original domains.
    \item We propose to automatically select the domain-specific predictor for an image during inference based on the minimum distance to mean features of each domain.
    \item We perform extensive experiments to demonstrate a promising balance between the stability (remembering old knowledge) and the plasticity (acquiring new knowledge) of the proposed method.
\end{itemize}

The remainder of this paper is organized as follows. In Sec.~\ref{sec_related_work}, we discuss the necessary background and related studies. We present the proposed lifelong depth learning framework in Sec.~\ref{sec_method}.  We then provide extensive numerical evaluations in Sec.~\ref{sec_result} and finally conclude our work in Sec.~\ref{conclusion}.

\section{Related Works}
\label{sec_related_work}

\begin{table*}[t]
\caption{Comparisons between several representative existing works and our work.}
\renewcommand\arraystretch{1.2}
\begin{center}
\label{property}
\begin{tabular}
{ccccc}
\hline
Methods & Lifelong learning  &  Scale-aware   & Cross-domain learning strategy &(Un)supervised learning\\ 
\hline
Virtual Normal v1 \cite{Yin2019enforcing} &\ding{55} &\ding{51}  & \ding{55} & Supervised\\
Virtual Normal v2 \cite{yin2022towards} &\ding{55} &\ding{55}   & Mixed data &Supervised\\ 
DABC \cite{li2018deep} &\ding{55} &\ding{51}   & Mixed data &Supervised\\ 
MiDas \cite{ranftl2020towards} &\ding{55} &\ding{55}  &Mixed data & Supervised\\ 
CoSelfDepth \cite{Khan2021TowardsCO} &\ding{51} &\ding{55}  & Mixed data &Unsupervised\\ 
Ours &\ding{51} &\ding{51}  &Sequential learning  & Supervised\\ 
\hline
\end{tabular}
\end{center}
\end{table*}

\subsection{Monocular Depth Estimation}
In recent years, monocular depth estimation has been formulated in a data-driven fashion either by penalizing pixel-wise loss between predicted depth maps and ground truth depth maps in supervised learning \cite{Wofk2019FastDepthFM,laina2016deeper,fu2018deep,Liu2018CollaborativeDN,Hu2019RevisitingSI,huynh2020guiding} or complying with the geometry consistency of multi-views in unsupervised learning \cite{Zhou2017UnsupervisedLO,song2021self,Zhan2018UnsupervisedLO,jiang2021plnet,Zhao2021MaskedGF,Sun2022UnsupervisedEO}. The advantage of unsupervised approaches is they can learn from videos and thus are easy to implement. However, their greatest drawback is that they only estimate relative depth maps and are highly limited for many applications, \eg, robot navigation.

While there has been significant progress in learning specific data domains, previous methods face severe challenges when deploying in real-world applications due to their poor performance in robustness and generalization.  One effective approach is to address these challenges by collecting a large-scale dataset encompassing diverse domains and training a domain-invariant model, as demonstrated in recent research works such as \cite{ranftl2020towards, xian2018monocular, ranftl2021vision, tro-depth, yin2022towards}. This solution is rather straightforward and still costly. Furthermore, if some new target domains appear, the model has to be learned from scratch. Therefore, there is an urgent demand to develop lifelong learning models as human vision systems.

In this paper, we study a lifelong learning paradigm that enables extending a single model for MDE to multiple domains sequentially.  We list the comparisons between this work and existing works in Table~\ref{property}. Compared to other methods of multi-domain learning using mixed data training strategy, we only reuse a few training data (less than $1\%$) from each of the old domains. A few works also studied LL for unsupervised MDE. by using replay data \cite{kuznietsov2021comoda,Khan2021TowardsCO} or additional parameters for domain adaptation \cite{zhang2020online}. They perform cross-domain learning with a small domain gap \cite{kuznietsov2021comoda,zhang2020online} or pre-training with mixed domain data \cite{Khan2021TowardsCO}. We differ from them in three major aspects: i) we can perform metric (absolute) depth estimation, which is more challenging than those methods in LL, ii) we can perform cross-domain learning with significant domain gaps, and iii) we can perform long-range lifelong learning across three different domains while previous works only successfully conducted lifelong learning between two similar domains.

\subsection{Lifelong Learning}
Lifelong learning (LL), also called incremental learning or continual learning, has been an active topic in machine learning. It aims to enable  ifelong learning of a model on new concepts/tasks/domains while preventing forgetting the previously learned knowledge. Most existing works solve LL on image recognition, and various approaches have been proposed. In general, those methods can be categorized into three types \cite{de2021continual}: i) replay methods \cite{rebuffi2017icarl,hou2019learning} that store some training samples for each of the previous tasks and reuse them while learning a new task, ii) regularization-based methods that prevents forgetting by imposing extra regularization instead of storing training samples, such as LwF \cite{li2017learning} using a knowledge distillation loss on previous tasks or EWC \cite{kirkpatrick2017overcoming} enforcing an additional loss term to alleviate changing on the weights important for previous tasks,  and iii) parameter isolation methods that fix trained parameters on old tasks and employ extra network branches for training a new task \cite{serra2018overcoming,rusu2016progressive}. It is worth mentioning that ExpertGate \cite{aljundi2017expert} assumes there are multiple expert models corresponding to multi-tasks. It proposed to learn domain-specific auto-decoder for each domain and use the minimum image reconstruction error to select the corresponding expert model. This method is straightforward and memory inefficient. For parameter isolation methods, the minimum increase of parameters is expected.

However, it is largely unknown how to enable LL for dense regression tasks, such as MDE. This work aims to disclose the difficulties and provide solutions for LL on MDE. To minimize the use of historical data and keep the increase in parameters for new domains as minimal as possible, our method enables lifelong learning through regularization by introducing uncertainty-aware knowledge preservation. Besides, as mentioned before, we also use a few replay data to mitigate significant domain gaps further. The replay data is also utilized for online domain identification during inference.

\begin{figure*}[t!]
    \centering
    \includegraphics[width = 0.96\linewidth]{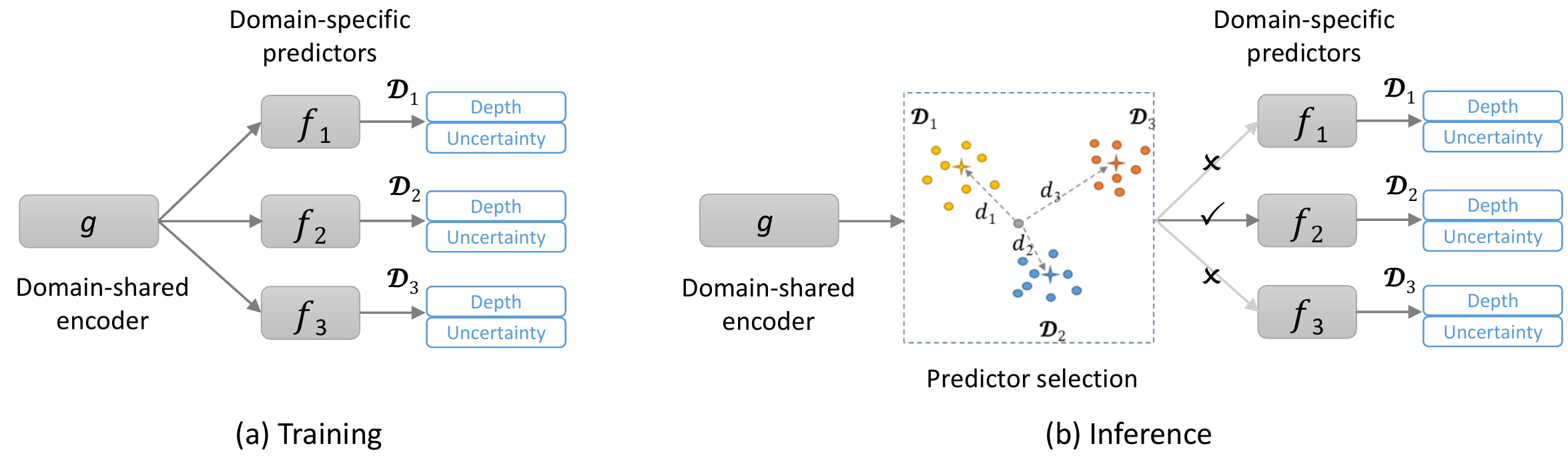}
    \caption{The diagram of the proposed lifelong learning framework for multi-domain metric depth estimation. $\mathcal{D}_i$ denotes $i$-th domain/task and $f_i$ denotes the corresponding predictor for scale-aware depth estimation. }
    \label{fig-method}
\end{figure*}

\section{Method}
\label{sec_method}
\subsection{Multi-head Depth Prediction Framework }
We follow most previous works in a practical setting that assumes training data of previous tasks are unavailable when learning a new task. 
Since the scene scale of depth is domain-dependent, we design a framework with multi-head depth predictors for domain-specific inference and a shared encoder for feature extraction. Each predictor is learned to estimate depth maps of a specific domain with a fixed depth range. A visualization of our framework is given in Fig.~\ref{fig-method} (a), where the model is shown for learning three different domains. 
The model starts from one depth predictor, \ie, $f_1$, for learning on $\mathcal{D}_1$ and extends its predictors dynamically and sequentially for learning on domains $\mathcal{D}_2$ and $\mathcal{D}_3$.  

\begin{table}[t]
\caption{Number of parameters of the feature extractor and domain-specific predictors.}
\renewcommand\arraystretch{1.2}
\begin{center}
\label{results_params}
\begin{tabular}
{c|c}
\hline
Module & Parameters (M)  \\
\hline
$g$ & 21.81 \\
$f_1$ &0.21 \\
$f_2$ &0.21\\
$f_3$ &0.21\\
\hline
\end{tabular}
\end{center}
\end{table}

Then, the problem is that such multi-head architecture will significantly increase model complexity for learning on multi-domains. To alleviate this issue, we introduce an extremely lightweight design using only two convolutional layers for depth prediction in each predictor. We take a compact pyramid feature network \cite{hu2021boosting} in which features extracted at each scale from the encoder are fused and compressed to a small fixed number; then, they are concatenated and inputted to the predictors. 
In our framework, the shared encoder is built on a ResNet-34 \cite{He2016DeepRL} and has 21.81 M parameters; each domain-specific predictor has 0.21 M parameters, respectively. In the case of learning three domains, the framework yields over 97\% shared parameters to promote computational efficiency. A more detailed description of the framework is given in the supplementary file.

During inference, the framework needs to identify the correct predictor for an input image. We propose an efficient method for this purpose.

\subsection{Lifelong Cross-Domain Depth Learning}
\subsubsection{Uncertainty-aware Knowledge Acquisition}
Given a target domain $\mathcal{D}^t$, where $\mathcal{D}^t = \{x^t, y^t\}$, $x^t$ and $y^t$ denote images and their corresponding depth maps, we can directly let the model learn to estimate depth maps in the target domain. 

Depth maps captured by sensors are usually sparse, suffer from outliers, and miss valid information around object boundaries, as seen in Fig.~\ref{fig-example}.
To eliminate the effect of outliers and improve the robustness, we employ an uncertainty-aware loss \cite{Hu2022DeepDC} as follows:
\begin{equation}
    \ell_{ud} = \sum (e^{-s^t}(\hat{y}^t-y^t)^{2} + s^t),
 \label{un_loss}
\end{equation}
where $\hat{y}^t$ is predicted depth maps from $x^t$, $e$ denotes the exponential operation, $s^t$ denotes pixel-wise uncertainty maps estimated simultaneously with depth maps, such that:
\begin{equation}
    \hat{y}^t, s^t = f_t(g(x^t)).
    \label{infererence}
\end{equation}

Similar to the depth estimation layers, we also use the two convolutional layers for uncertainty estimation, resulting in a total of 0.21 M parameters for each domain-specific predictor. Besides, the uncertainty estimation layers can be dropped during inference for efficient computation.

\subsubsection{Uncertainty-aware Knowledge Preservation}
For learning on a new domain $\mathcal{D}^{t+1}$, we accordingly add a new domain-specific depth predictor $f_{t+1}$ such that $\hat{y}^{t+1}=f_{t+1}(g(x^{t+1}))$ and learn its parameters with Eq.~\eqref{un_loss}. However, this will shift the parameters of the encoder, thus leading the estimation on $\mathcal{D}^{1,2,...,t}$ to malfunction, \ie, causing catastrophic forgetting.

As studied in \cite{hu2022data}, applying KD with out-of-distribution data is able to distill the knowledge of a model learned on the original domain for MDE.
In our method, the trained model on $\mathcal{D}^1,...,\mathcal{D}^t$ serves as an expert teacher and provides desired predictions on each domain. 
Formally, we let $g$ and $f_1,...,f_{t+1}$ denote the new encoder and domain-specific predictors while performing lifelong learning on $\mathcal{D}^{t+1}$, let $g'$ and $f'_1,...,f'_t$ be the old model learned on $\mathcal{D}^1,...,\mathcal{D}^t$.
We apply regularization on both depth consistency and uncertainty consistency on $\mathcal{D}^{i}, i \in \{1,2,...,t\} $ as follows:
 \begin{equation}
 \begin{aligned}
   &  \ell_{cons} = \sum (|\hat{y}^i_n - \hat{y}^i_o| + |s^i_n - s^i_o|), \\
     & \text{s.t.} \ \ \    \hat{y}^i_n, s^i_n = f_i(g(x^{t+1})) \\
      & \qquad \hat{y}^i_o, s^i_o = f'_i(g'(x^{t+1})), \\
    \label{kd_loss}
 \end{aligned}
\end{equation}
where $\hat{y}^i_n$ and $\hat{y}^i_o$ denotes predicted depth images of $\mathcal{D}^{i}$ with the new model and old model, respectively; similarly,  $s^i_n$ and $s^i_o$ are predicted uncertainty with new model and old model.

\subsubsection{Replay for Memory Enhancement}
If images from $D^{t+1}$ lie in the same distribution as images from $D^{i}$, \ie, $\mathcal{P}(x^i)==\mathcal{P}(x^{t+1})$, Eq.\eqref{kd_loss} will be fully effective for preserving knowledge on $\mathcal{D}^i$. Otherwise, its performance tends to degrade due to the domain gap. 
Therefore, there is a risk that the model will significantly deteriorate its performance on previous domains because of a significant domain shift between $\mathcal{D}^{t+1}$ and $D^i$ where $i \in \{1,2,...,t\}$.

To handle this issue, we take a replay strategy as many classical lifelong learning methods \cite{rebuffi2017icarl,hou2019learning}, which is more consistent with human cognition by periodically and repeatedly reviewing historical data. We randomly
preserve limited training data (500 images) of each of the previous domains and replay them for learning on new domains. Then, the replay loss is formulated as:
\begin{equation}
      \ell_{replay} =  \ell_{ud}(\hat{y}^i, y^i).
 \label{replay_loss}
\end{equation}

Then, the loss for lifelong learning on $\mathcal{D}^{t+1}$ can be written as:
\begin{equation}
\begin{gathered}
      \mathcal{L} = \sum_{i=1}^{t}  \lambda^i \left( \ell_{cons}(\hat{y}^i_n, s^i_n, \hat{y}^i_o, s^i_o) + \ell_{replay}(\hat{y}^i, y^i) \right) \\ +  \ell_{ud}(\hat{y}^{t+1}, y^{t+1}), 
 \label{total_loss}
 \end{gathered}
\end{equation}
 where $\lambda$ is a vector and $\lambda^t$ denotes the weight coefficient for domain $\mathcal{D}^t$.
The first and the second loss term in Eq.\eqref{total_loss} alleviate knowledge forgetting on domains $\mathcal{D}^1$ to $\mathcal{D}^t$, the third loss term in Eq.\eqref{total_loss} promotes learning knowledge on the new target domain $\mathcal{D}^{t+1}$. 

\begin{algorithm}[t]  
  \caption{Lifelong-MonoDepth: Training}  
  \label{alg_training}  
  \begin{algorithmic}[1]  
    \Require  
    
      $\mathcal{D}^{t+1}$: new target domain; 
      
      $N^t = \{g', f'_1, ..., f'_t\}$: old model; 
      
      $\lambda^t$: weight coefficients; 
      
      $\mathcal{P}=\{\mathcal{P}_1,...,\mathcal{P}_t\}$: replay sets; 
    \Ensure  $N^{t+1} =\{g, f_1, ..., f_{t+1}\}$: new model;
        \State Freeze $N^t$;
        \For{$j$ = 1 to $iterations$}             
            \Statex \LeftComment{1} {\color{gray} \% knowledge acquisition from new domain \%}   
            \State Set gradients of $N^{t+1}$ to 0;
            \State Select a batch $(x^{t+1},y^{t+1})$ from $\mathcal{D}^{t+1}$;
            \State Get predictions $\hat{y}^{t=1},s^{t+1}$  $\leftarrow$ \ $f_{t+1}(g(x^{t+1}))$;
            \State Compute uncertainty-aware depth loss $\ell_{ud}$ by Eq.\eqref{un_loss};
            \Statex \LeftComment{1} {\color{gray} \% knowledge preservation for old domains \%}
            \For{$i$ = 1 to $t$}             
                \State Get consistency loss $\ell_{cons}$ by Eq.\eqref{kd_loss};
                \State Select a batch $(x^{i},y^{i})$ from $\mathcal{P}_i$;
                \State Compute replay loss $\ell_{replay}$ by Eq.\eqref{replay_loss};                 
            \EndFor
        \State Get the total loss $\mathcal{L} = \ell_{ud} + \lambda^i  \sum_{i=1}^{t} (\ell_{cons} +   \ell_{replay})$;
        \State Backpropagate $\mathcal{L}$;
        \State Update $N^{t+1}$; 
        \EndFor           
  \end{algorithmic}  
\end{algorithm} 

\begin{algorithm}[t]  
  \caption{Lifelong-MonoDepth: Inference}  
  \label{alg_inference}  
  \begin{algorithmic}[1]  
    \Require  
    
      $N^t = \{g, f_1, ..., f_{t}\}$: learned model on  $\mathcal{D}^1$ to $\mathcal{D}^t$; 
      
      $\mu=\{\mu_1,...,\mu_t\}$: domain-specific mean features; 

      $x$: an image from any domain $\mathcal{D}^i, i\in\{1,...,t\}$;
  
    \Ensure  $\hat{y}$: a depth map;
        \State Compute intermediate features by $g(x)$;
        \For{$i$ = 1 to $t$}             
            \State Compute the distance $d_i$ between $g(x)$ and $\mu_i$;             
        \EndFor
        \State Select predictor $f_i \leftarrow \mathop{\arg\min}\limits d_i$;
       \State  Output depth map $\hat{y} \leftarrow f_i(g(x))$;
  \end{algorithmic}  
\end{algorithm}

\subsection{Online Cross-Domain Depth Inference}
After incremental learning on $\mathcal{D}^1$ to $\mathcal{D}^t$, ideally, the model is able to correctly estimate a depth map $\hat{y}$ from any image $x$ sampled from $D^i$, $i \in \{1,2,...,t\}$. A practical challenge is how to identify the domain of $x^i$ and accordingly select the corresponding predictor $f_i$ automatically during inference.

To address this problem, we propose to identify the minimum distance between a given image and each domain in the feature space. Since we preserve a small subset of each domain, we can obtain the mean features of each domain approximated with these replay data, that is:
\begin{equation}
      \mu^i =  \sum_{k=1}^{k} g(x^i_k),
\end{equation} 
where $x_k^i$ is $k$-th image of the replay set of the domain $D^i$,  $\mu^i$ is the mean features of $D^i$ calculated by the replay set. Then, identifying the domain to which the input image belongs can be formulated as:
\begin{equation}
 \begin{aligned}
    &  i  \leftarrow    \mathop{\arg\min}\limits_{i} d_i, \\
        & \text{s.t.} \ \ \  d_i = \| g(x) - u^i \|_2,\\
\end{aligned}
\label{find_f}
\end{equation} 
where $\| \cdot \|_2$ denotes the $\ell_2$ norm. Finally, the $i$-th predictor, i.e., $f_i$ corresponding to the domain $\mathcal{D}^i$ is selected for depth inference. 

 \begin{table}[t]
\caption{Details of the RGBD datasets used in the experiments.}
\renewcommand\arraystretch{1.2}
\begin{center}
\label{datasets}
\begin{tabular}
{c|ccc}
\hline
\multirow{2}{*}{Dataset} &\multirow{2}{*}{Depth range (m)} & Training & Test \\ 
 && scenarios / images & scenarios / images \\ \hline
 NYU-v2 &0 $\sim$ 10 &249 / 50688 & 215 / 654\\
 KITTI  & 0 $\sim$ 80 & 138 / 85898 & 18 / 1000\\ 
 ScanNet &0 $\sim$ 6 &1513 / 50473 &100 / 17607\\ 
\hline
\end{tabular}
\end{center}
\end{table}

\section{Experiments}
\label{sec_result}
\subsection{Experimental Setup}
\subsubsection{Datasets}
We evaluate our method on three benchmark datasets, including two indoor and one outdoor dataset. The details are given as follows.

\paragraph{NYU-v2 \cite{NYUv2}} The NYU-v2 dataset is one of the most commonly used benchmarks for indoor depth estimation. NYU-v2 has 464 indoor scenes captured by Microsoft Kinect with an original resolution of $640 \times 480$. Among them, 249 scenes are used for training, and the rest 215 scenes are used for testing.
We use the pre-processed data by \cite{Hu2019RevisitingSI,hu2019analysis}
with about 50,000 RGBD pairs. 
Following previous studies, we resize the images to 320$\times$240 pixels and then crop their central parts of 304$\times$228 pixels as inputs. 
For testing, we use the official small subset of 654 RGBD pairs.

\begin{table*}[t!]
\caption{Quantitative comparisons between existing methods and the proposed method in which $\&$ denotes data mixing and $\rightarrow$ denotes sequential order for lifelong learning. Note that we specify the correct domain-specific predictor for each input image. $*$ denotes results taken from \cite{Khan2021TowardsCO}.}
\renewcommand\arraystretch{1.2}
\begin{center}
\label{results1}
\begin{tabular}
{c|ccc|ccc|ccc}
\hline
&\multicolumn{3}{c|}{ NYU-v2} &\multicolumn{3}{c|}{ KITTI} &\multicolumn{3}{c}{ Average}  \\ 
Method & RMSE &REL  &$\delta_1$ & RMSE &REL &$\delta_1$  & RMSE &REL &$\delta_1$   \\ \hline
SDT &0.532 &0.130  &0.836 &3.286 &0.070 &0.939 &1.909 &0.100 &0.888\\ 
JDT (NYU-v2 $\&$ KITTI) &0.581 &0.151 &0.803 &3.658 &0.086 &0.911 &2.120 &0.119 &0.857\\ 
\hline
Comoda$^*$ \cite{kuznietsov2021comoda} (NYU-v2 $\&$ KITTI $\rightarrow$ KITTI)  &0.673 &0.191 &0.706 &6.249 &0.158 &0.769 &3.461 &0.175 &0.738\\
CoSelfDepth$^*$ \cite{Khan2021TowardsCO}  (NYU-v2 $\&$ KITTI $\rightarrow$ KITTI)  &0.626 &0.187 & 0.728 &5.809 &0.154 &0.784 &3.218 &0.171 &0.756\\
\hline
FT (NYU-v2 $\rightarrow$ KITTI) &1.133 &0.328 &0.451  &3.655 &0.079 &0.918 &2.394 &0.204 &0.685\\ 
FAL (NYU-v2 $\rightarrow$ KITTI) & 0.532 &0.130  &0.836 &8.946 &0.252 &0.600 &4.739 &0.191 &0.718 \\
EWC (NYU-v2 $\rightarrow$ KITTI) &1.007 &0.251 &0.475  &4.550 &0.100  &0.876 &2.779 &0.176 &0.676\\ 
Ours (NYU-v2 $\rightarrow$ KITTI) &0.622 &0.162 &0.768  &3.829 &0.081 &0.910 &\textbf{2.226} &\textbf{0.122} &\textbf{0.839}\\ 
\hline
FT (KITTI $\rightarrow$ NYU-v2) &0.555 &0.137 &0.820  &13.22 &0.450  &0.179 &6.888 &0.294 &0.500\\ 
FAL (KITTI $\rightarrow$ NYU-v2) &0.991 &0.318 & 0.523  &3.286 &0.070  &0.939 &2.139 &0.194 &0.731\\ 

EWC (KITTI $\rightarrow$ NYU-v2) &0.650 &0.173  &0.755  &7.178 &0.243  &0.573 &3.914 &0.208 &0.664\\ 

Ours (KITTI $\rightarrow$ NYU-v2) &0.567 &0.142 &0.812  &5.060 & 0.136  &0.813 &\textbf{2.814} & \textbf{0.139} &\textbf{0.813}\\ 
\hline
\end{tabular}
\end{center}
\end{table*}

\paragraph{KITTI \cite{SICNN}: }  
This outdoor dataset, collected by car-mounted cameras and a LIDAR sensor, was also widely used as a benchmark in previous studies of MDE. We use the official KITTI depth prediction dataset with
the official split of scenes for training and validation. The training and validation set has 138 and 18 driving sequences, respectively.
The resolution is about $1216 \times 352$ for most images. We randomly crop a patch with $480 \times 320 $ resolution for training and use the original resolution for testing.

 \paragraph{ScanNet \cite{dai2017scannet}} ScanNet is a large-scale indoor RGBD dataset that contains 2.5 million RGBD images. We randomly and uniformly select a subset of approximately 50,000 samples from the training splits of 1513 scenes for training and evaluate the models on the test set of another 100 scenes with 17K RGB pairs. The resolution of RGB images is 1296 $\times$ 968.
 We apply image resizing and cropping as utilized on the NYU-v2 dataset.

\subsubsection{Implementation Details}
We train the model for 20 epochs using the Adam optimizer with an initial learning rate of 0.0001 for each dataset and reduce it to 50\% for every five epochs. While learning on $\mathcal{D}^{t+1}$,
the hyper-parameters $\lambda^i$ for preventing forgetting on $\mathcal{D}^i$ are set to 10 for the indoor dataset and 100 for the outdoor dataset for all experiments throughout the paper.
 We trained models with a batch size of 8 in 
all the experiments using PyTorch \cite{NEURIPS2019_9015}. 
For the sake of fair comparison, we train with the uncertainty-aware loss function for all baseline methods. 
Notably, as the depth scale is significantly different across domains, as seen in Table~\ref{datasets}, we apply a scale-invariant operation to depth maps in the loss function by dividing the median depth value of ground truth to exclude potential disturbance.

For evaluation, we use the most popular three measures, including RMSE, REL, and $\delta_1$. The first is a scale-aware measure, and the latter two are scale-invariant.

\begin{figure*}[t!]
\centering  
\begin{tabular}
{p{0.09\textwidth}<{\raggedright}p{0.09\textwidth}<{\centering}p{0.09\textwidth}<{\centering}p{0.26\textwidth}<{\centering}p{0.25\textwidth}<{\centering}} 
(a) Images
&\IncG [height=0.08\textwidth]{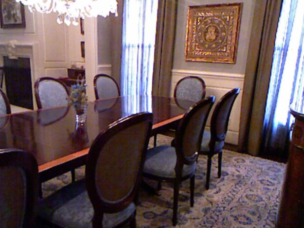} 
&\IncG [height=0.08\textwidth]{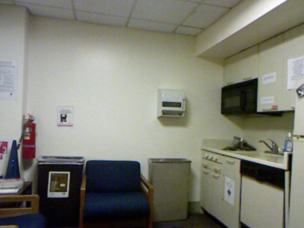} 
&\IncG [height=0.08\textwidth]{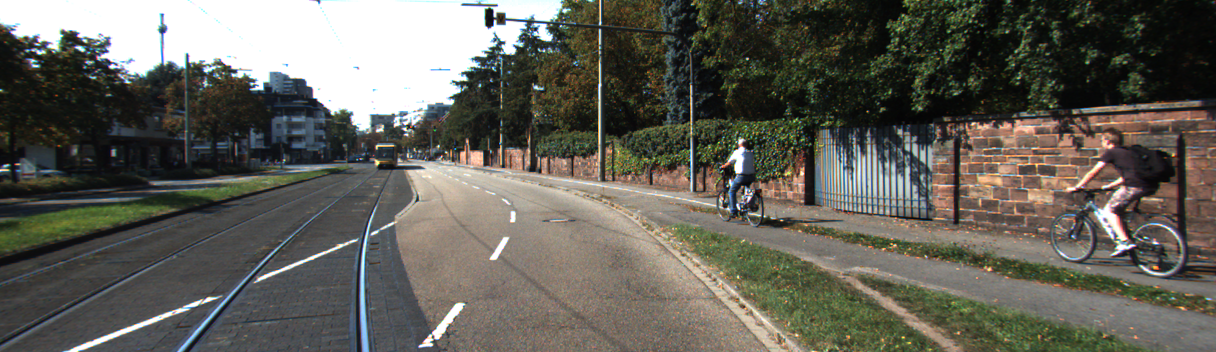} 
&\IncG [height=0.08\textwidth]{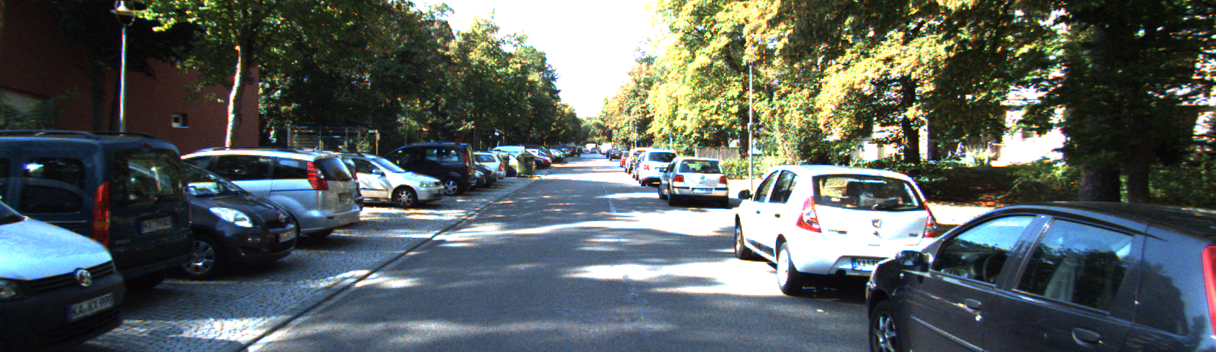} 
\\ 
(b) Ground Truths
&\IncG [height=0.08\textwidth]{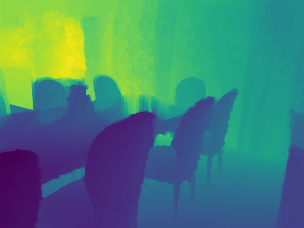} 
&\IncG [height=0.08\textwidth]{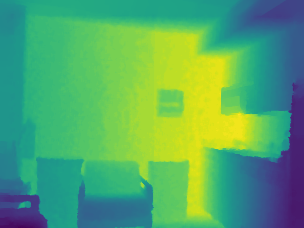} 
&\IncG [height=0.08\textwidth]{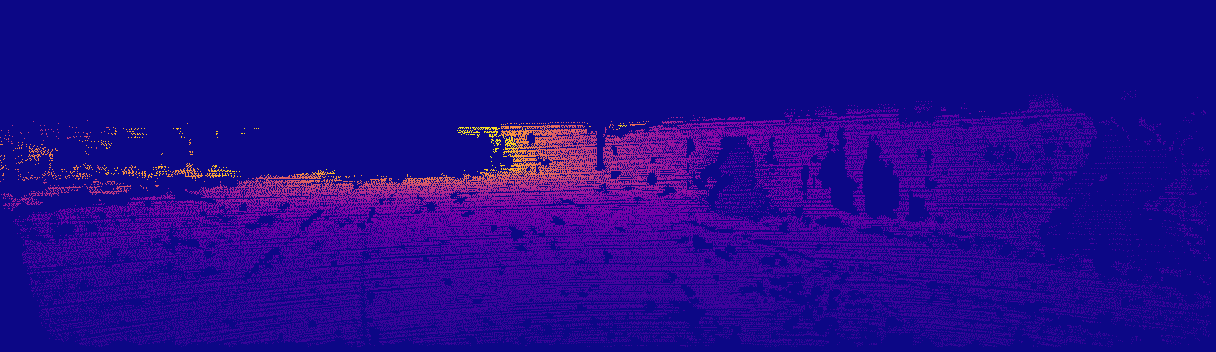} 
&\IncG [height=0.08\textwidth]{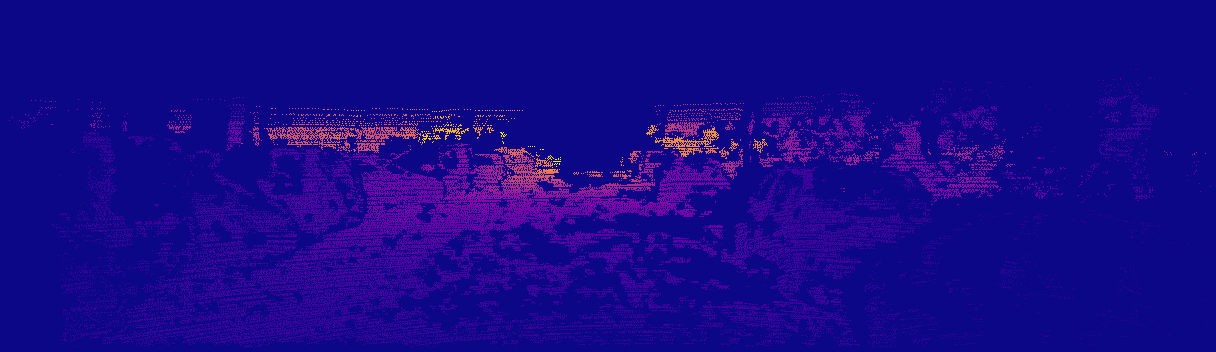} 
\\ 
(c) FT
&\IncG [height=0.08\textwidth]{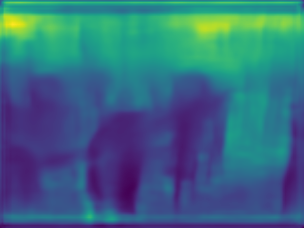} 
&\IncG [height=0.08\textwidth]{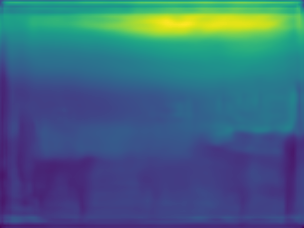} 
&\IncG [height=0.08\textwidth]{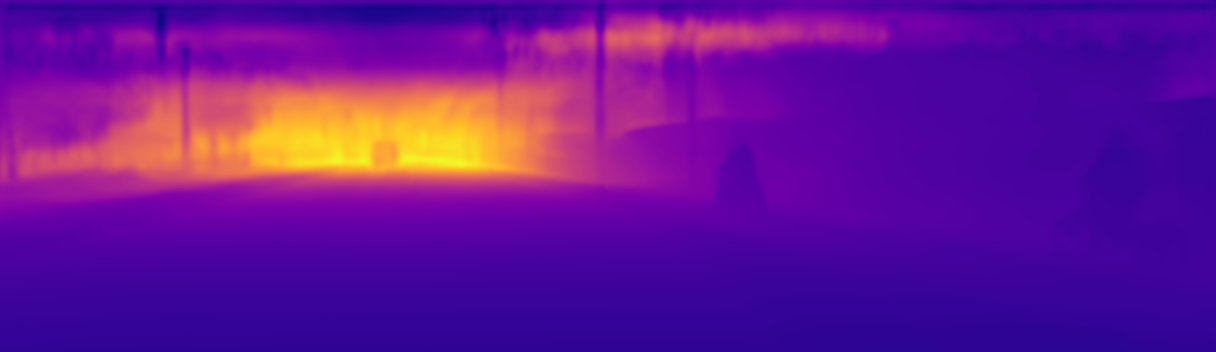} 
&\IncG [height=0.08\textwidth]{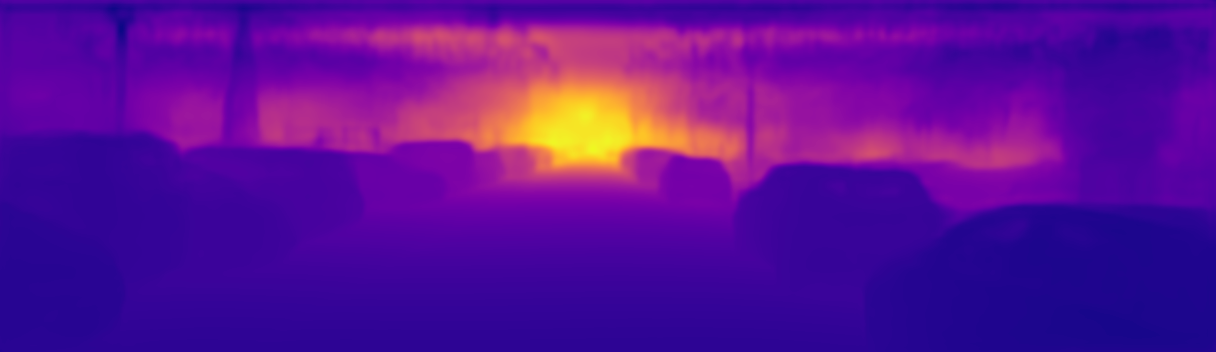} 
\\ 
(d) FAL
&\IncG [height=0.08\textwidth]{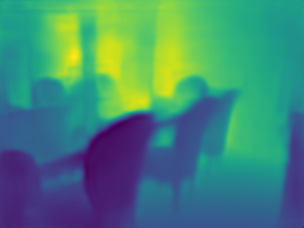} 
&\IncG [height=0.08\textwidth]{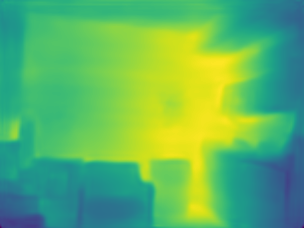} 
&\IncG [height=0.08\textwidth]{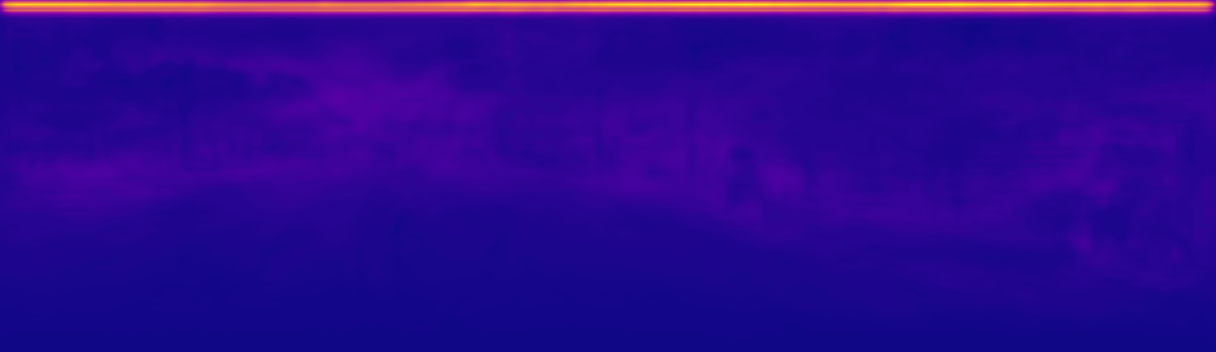} 
&\IncG [height=0.08\textwidth]{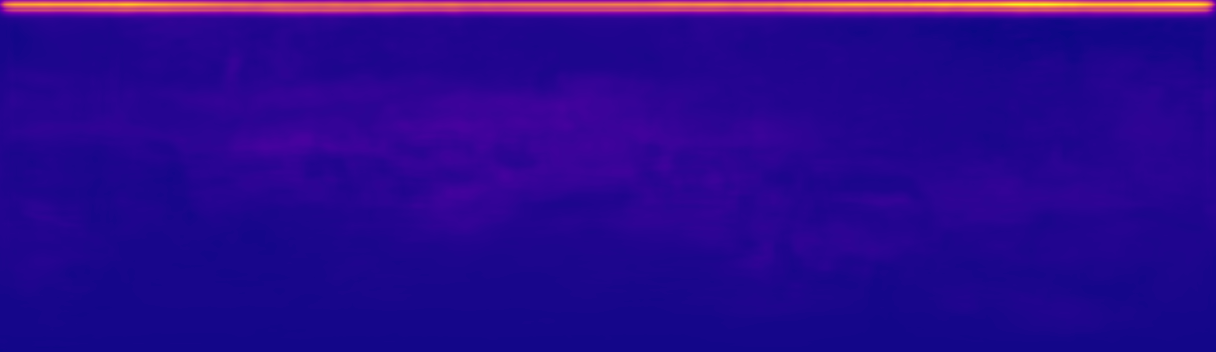} 
\\ 
(e) EWC
&\IncG [height=0.08\textwidth]{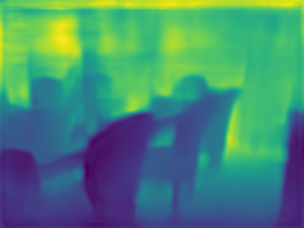} 
&\IncG [height=0.08\textwidth]{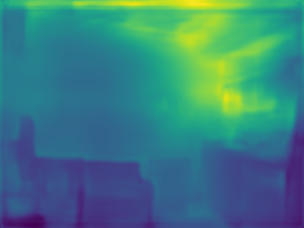} 
&\IncG [height=0.08\textwidth]{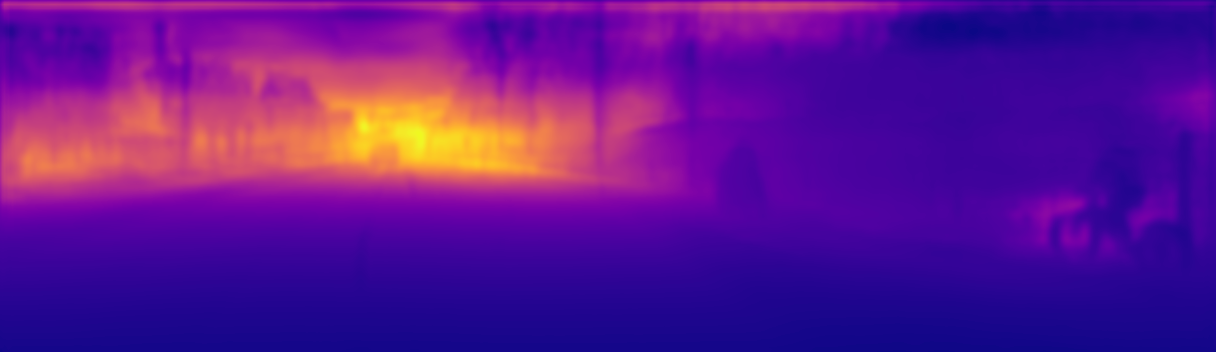} 
&\IncG [height=0.08\textwidth]{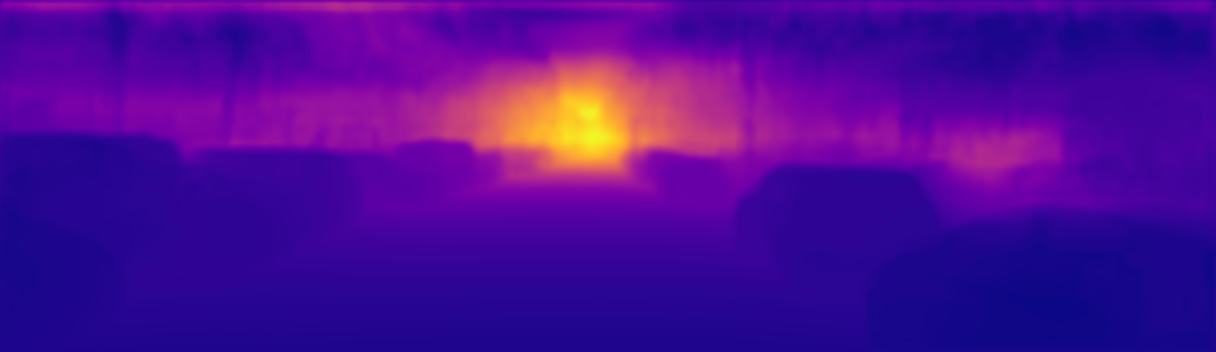} 
\\ 
(f) Ours
&\IncG [height=0.08\textwidth]{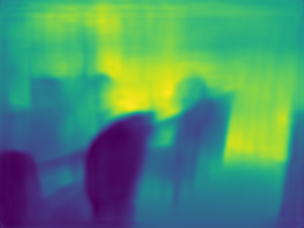} 
&\IncG [height=0.08\textwidth]{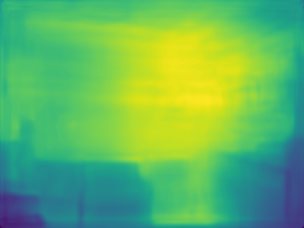} 
&\IncG [height=0.08\textwidth]{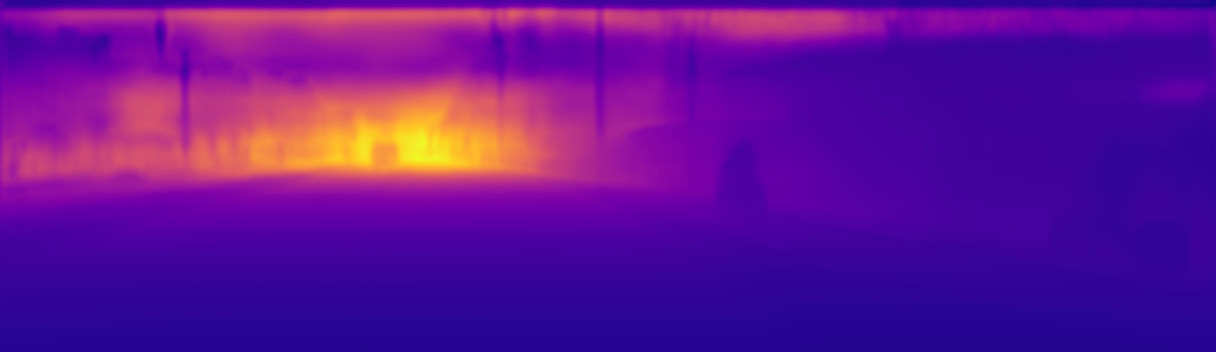} 
&\IncG [height=0.08\textwidth]{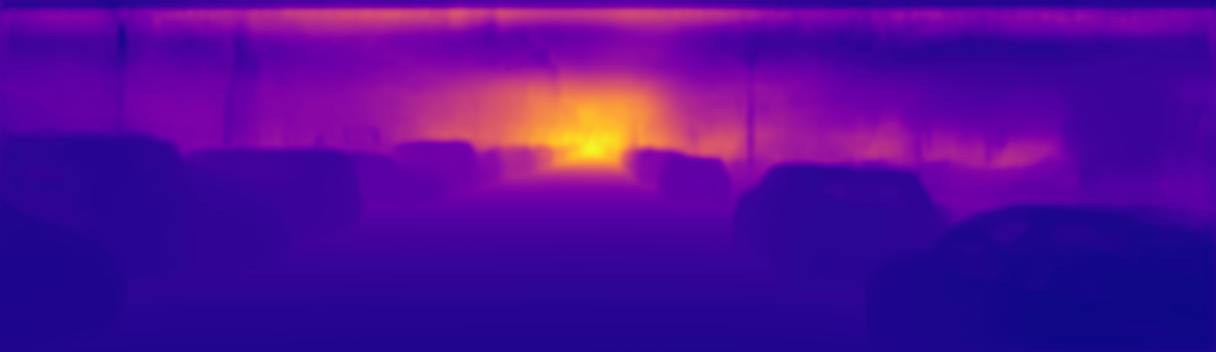} 
\\
\end{tabular}
\caption{Qualitative comparison of depth maps predicted by different methods in the learning order of NYU-v2 $\rightarrow$ KITTI.}
\label{fig-qua-res1}
\end{figure*}

\subsubsection{Baselines}
Since no previous methods have been proposed for lifelong metric depth learning, we consider different learning strategies as baselines for our method as follows.

\textbf{Single-Domain Training (SDT)}: is the standard learning protocol for single domain depth learning, \ie, training and evaluating on the same dataset. The performance of SDT provides an upper bound that we aim to reach.

\textbf{Joint-Domain Training (JDT)}: randomly selects a batch from each domain and then mixes data to perform joint learning.

\textbf{Fine-tuning (FT)}: is a common baseline for lifelong learning. We also compare it in experiments.

\textbf{Freezing And Learning (FAL)}: is a parameter isolation strategy. It
freezes old model parameters including $g, f_1, ..., f_t$, and only updates new model parameters, \ie, $f_{t+1}$ while learning on $\mathcal{D}^{t+1}$. 

\textbf{Elastic Weight Consolidation \cite{kirkpatrick2017overcoming} (EWC):} is a classical method for lifelong learning. It overcomes catastrophic forgetting by discouraging modifying weights important for old tasks.

Among baseline methods, \textbf{SDT} and \textbf{JDT} use a single-head network to demonstrate the upper bound of single-domain learning and multi-domain learning. The other methods adopt the same multi-head framework as our method.
We also compare two existing methods of lifelong depth learning, including
\textbf{Comoda \cite{kuznietsov2021comoda}} and
\textbf{CoSelfDepth \cite{Khan2021TowardsCO}}. These two methods are proposed for unsupervised depth learning and use data replay to avoid catastrophic forgetting. Note that the original Comoda targets outdoor autonomous driving scenes with small domain gaps and the code of CoSelfDepth is not publicly available. We take results implemented in CoSelfDepth \cite{Khan2021TowardsCO} for reference in which experiments on NYU-v2 and KITTI are performed.

\subsection{Results of Stability and Plasticity }
\subsubsection{Results on Two Domains}
We first conduct experiments on two domains, including NYU-v2 and KITTI, the domain gap between which is significant. We compare the proposed method against all baseline approaches. Since the learning order has a large impact on the results of each domain, we perform experiments in the order of both NYU-v2 $\rightarrow$ KITTI and KITTI $\rightarrow$ NYU-v2. The old domain is NYU-v2 and the new domain is KITTI for the former and reversely for the latter.
Notably, FT and FAL inherently yield the best plasticity and stability due to their training strategy.
Therefore, we also compute the average accuracy on the two datasets for better quantifying the trade-off between stability and plasticity.

The results are given in Table~\ref{results1}. No doubt, SDT and JDT gained the best and second-best performance, respectively. For results of cross-domain learning, although FT obtained the best accuracy on the new domain, it would yield extremely poor performance on the old domain, showing the worst result of mean accuracy. EWC can be seen as an improved method of FT that tackles this problem by employing an additional regularization term. We observe that EWC demonstrates slightly low performance than FT on the new domain, whereas it gained much better performance on the old domain, thus achieving better average accuracy.
In contrast, FAL does not suffer from catastrophic forgetting at the cost of sacrificing the plasticity on a new domain. As a result, our method achieves promising results for both the old and the new domain. Although it slightly underperforms FAL in stability showing the second-best performance on the old domain, we gained the best average accuracy for all three measures, \eg, outperforming FT, FAL, and EWC in $\delta_1$ by 15.4\%, 12.1\%, 13.6\% on NYU-v2 and 31.3\%, 8.2\%, 14.9\% on KITTI, respectively. 

The results of Comoda \cite{kuznietsov2021comoda} and CoSelfDepth \cite{Khan2021TowardsCO} are taken from \cite{Khan2021TowardsCO}. Note that the implementations are different from ours. Thus, we mark them in $*$. These two methods used half of the training data from NYU-v2 and KITTI for pre-training and then performed lifelong learning with the other half of the data. Hence, they suffer marginally from large domain shifts. Nevertheless, our approach demonstrates clearly better performance than the two methods.

Fig.~\ref{fig-qua-res1} shows qualitative comparisons between our method and baseline approaches. It is seen that FT predicted good depth maps on KTTI; however, it failed on NYU-v2. FAL inferred the best depth maps on NYU-v2, on the other hand, failed on KITTI. Both EWC and our method could produce perceptually correct depth maps and our method yield more accurate predictions on NYU-v2.

\begin{table}[t] 
\caption{The average accuracy of lifelong depth learning on all three domains with all six different learning orders.}
\renewcommand\arraystretch{1.2}
\begin{center}
\label{results_3domains}
\begin{tabular}
{c|ccc}
\hline
Learning order &RMSE & REL & $\delta_1$ \\ \hline
KITTI $\rightarrow$ NYU-v2 $\rightarrow$ ScanNet) &3.314 &0.181 &0.694 \\ 
KITTI $\rightarrow$ ScanNet $\rightarrow$ NYU-v2) &2.871 &0.173  &0.729\\ 
NYU-v2 $\rightarrow$ KITTI $\rightarrow$ ScanNet) &2.263 &0.145  &0.780  \\ 
ScanNet $\rightarrow$ KITTI $\rightarrow$ NYU-v2) &2.075 &0.146  &0.774\\ 
NYU-v2 $\rightarrow$ ScanNet $\rightarrow$ KITTI) &1.716 &0.145 &0.779\\ 
ScanNet $\rightarrow$ NYU-v2 $\rightarrow$ KITTI) &1.644 &0.146 &0.782\\ \hline
Average  &2.314 &0.156 &0.756\\ 
\hline
\end{tabular}
\end{center}
\end{table}

\begin{figure}[h]
\centering
\subfigure[\footnotesize{KITTI $\rightarrow$ NYU-v2 $\rightarrow$ ScanNet}]
{\includegraphics[width=0.22\textwidth]{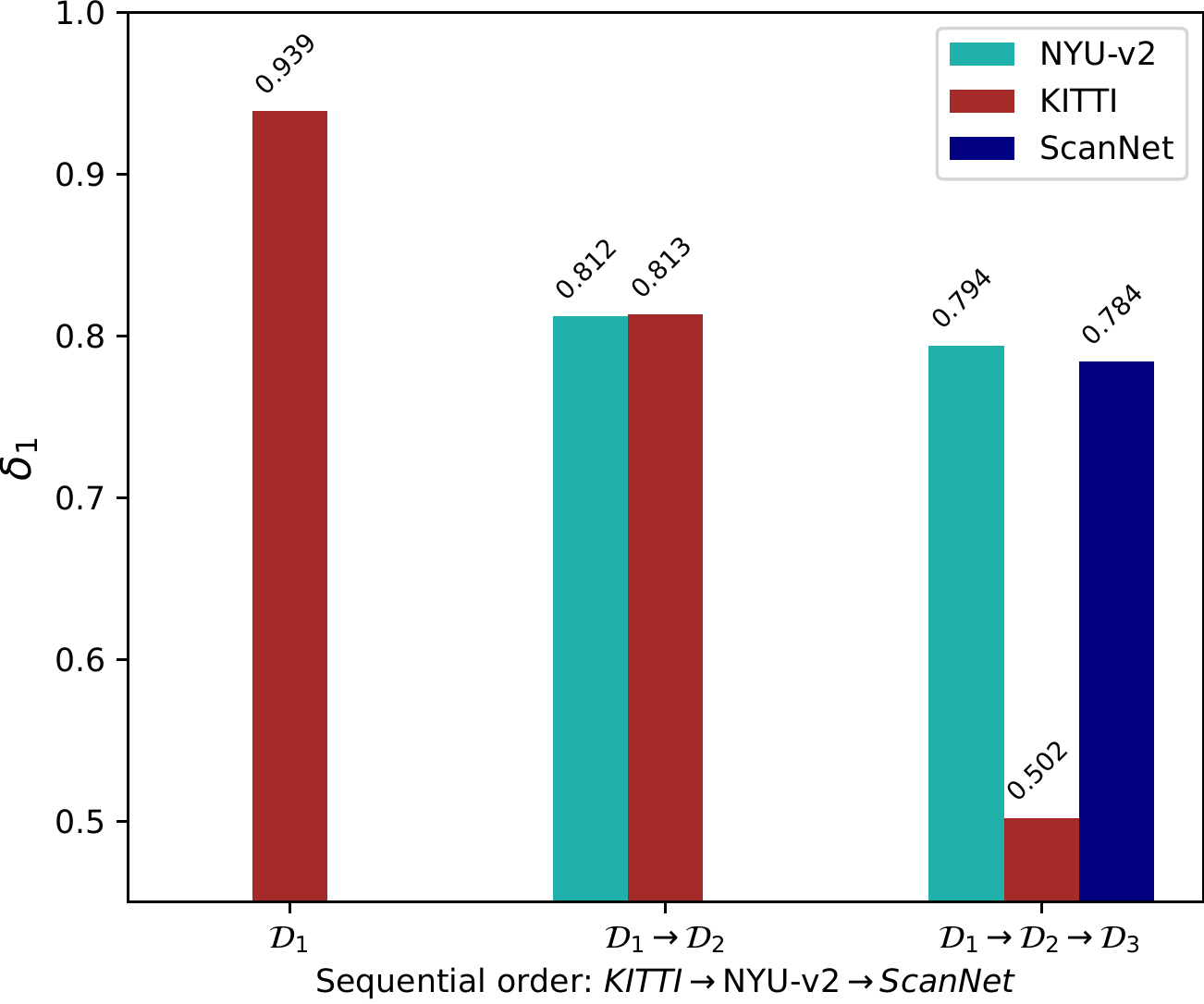}}
\subfigure[\footnotesize{KITTI $\rightarrow$ ScanNet $\rightarrow$ NYU-v2}]
{\includegraphics[width=0.22\textwidth]{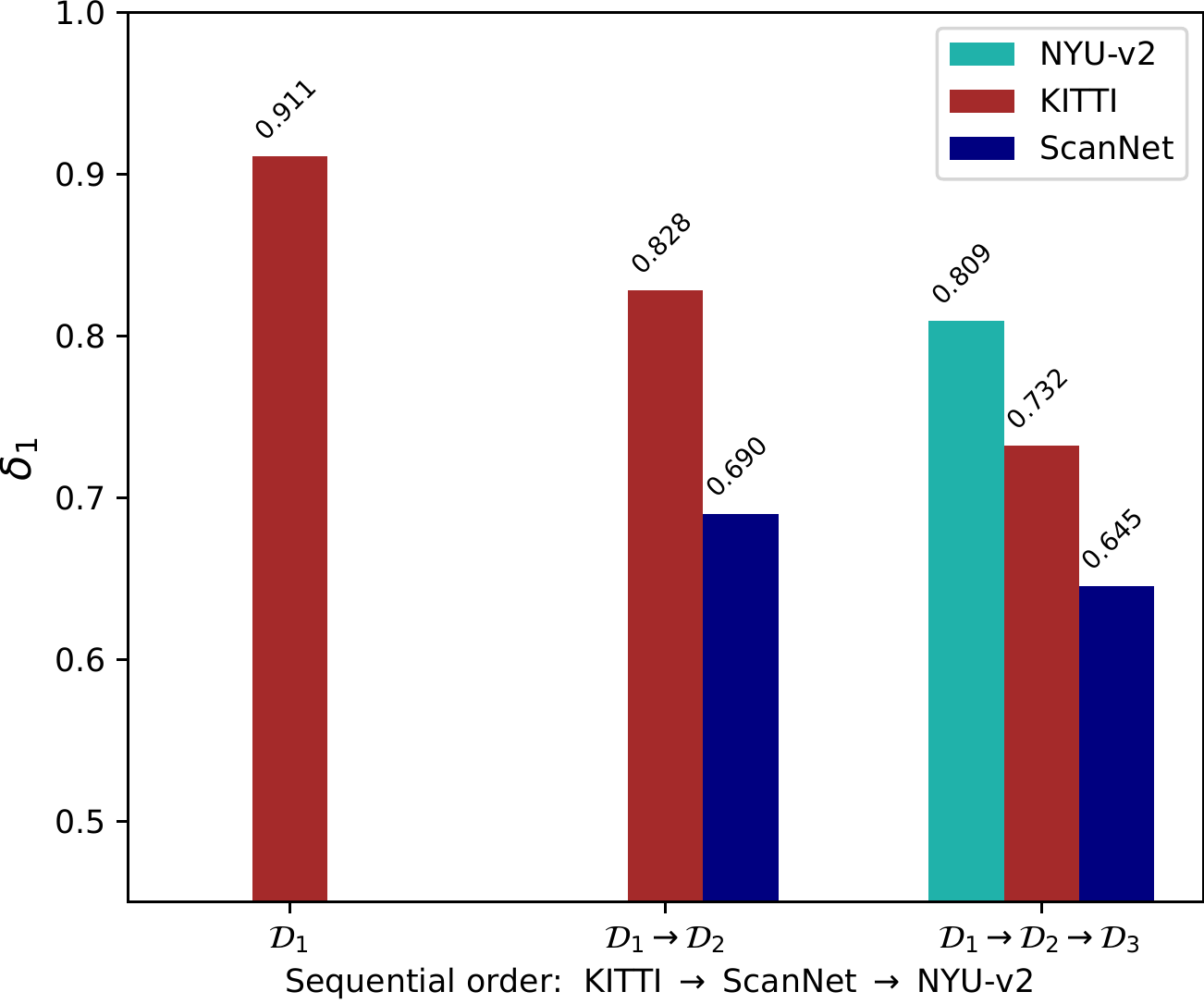}} \\
\subfigure[\footnotesize{NYU-v2 $\rightarrow$  KITTI  $\rightarrow$ ScanNet}]
{\includegraphics[width=0.22\textwidth]{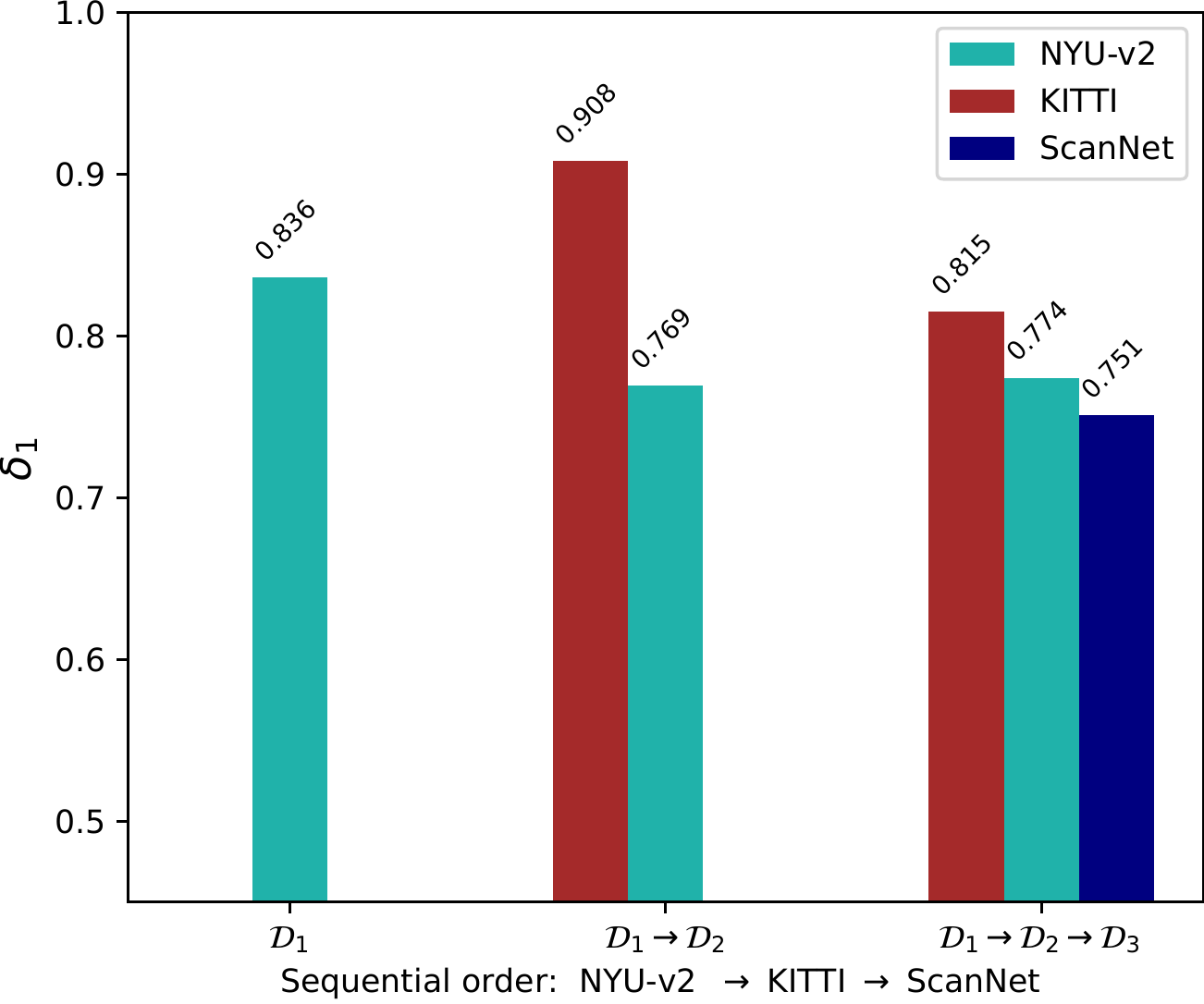}}
\subfigure[\footnotesize{ScanNet $\rightarrow$ KITTI $\rightarrow$ NYU-v2 }]
{\includegraphics[width=0.22\textwidth]{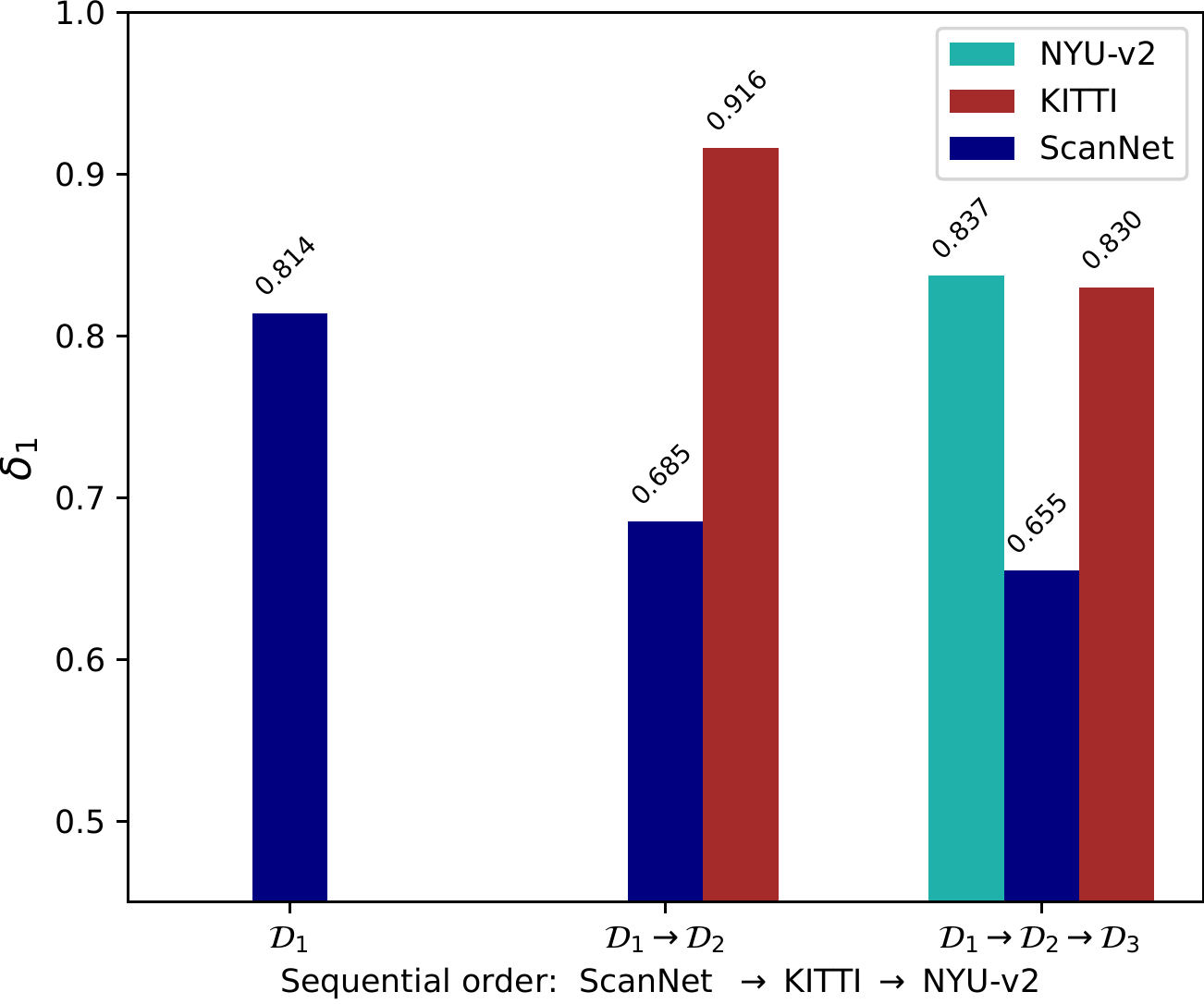}} \\
\subfigure[\footnotesize{NYU-v2 $\rightarrow$  ScanNet $\rightarrow$ KITTI }]
{\includegraphics[width=0.22\textwidth]{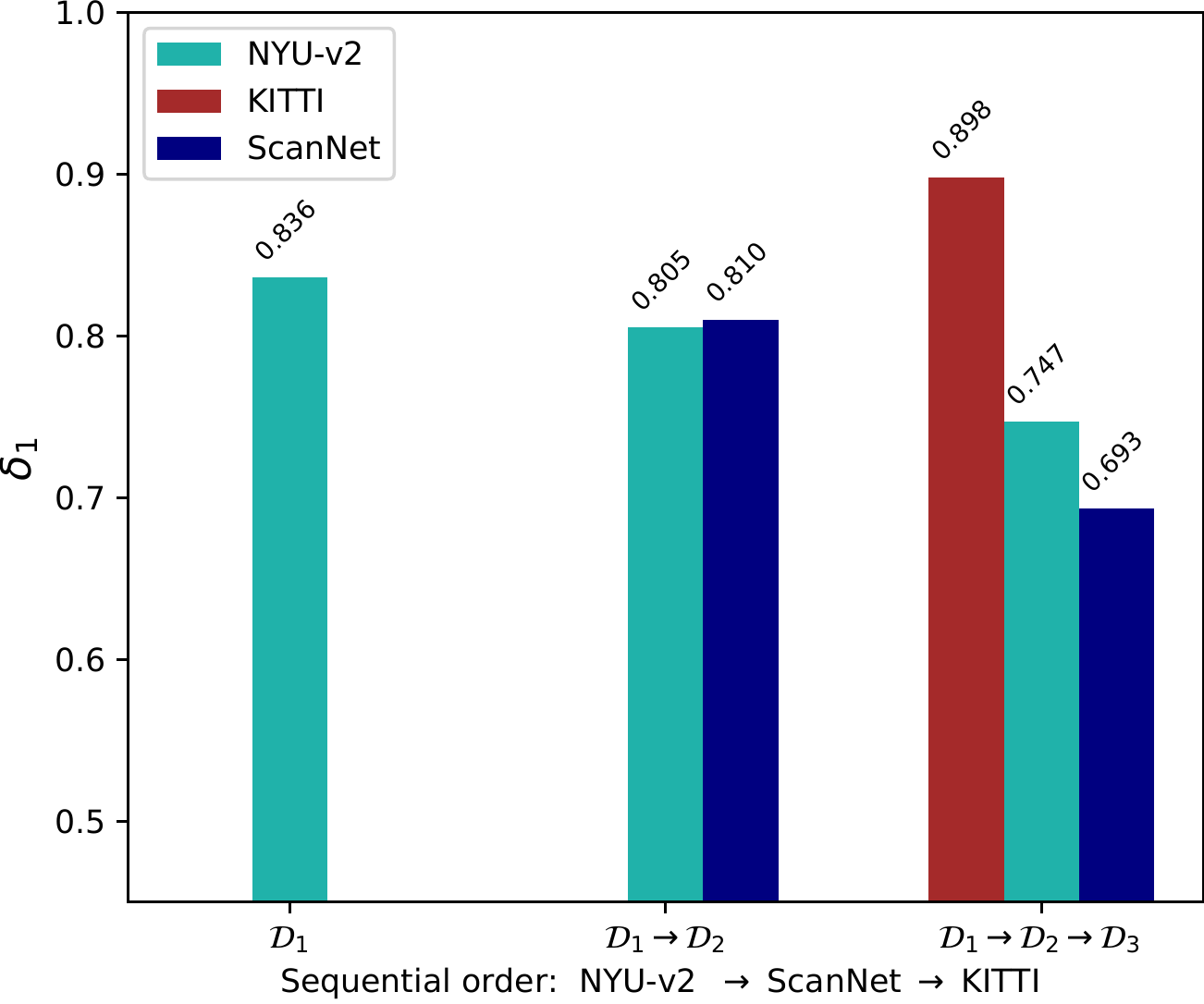}}
\subfigure[\footnotesize{ScanNet $\rightarrow$ NYU-v2 $\rightarrow$ KITTI}]
{\includegraphics[width=0.22\textwidth]{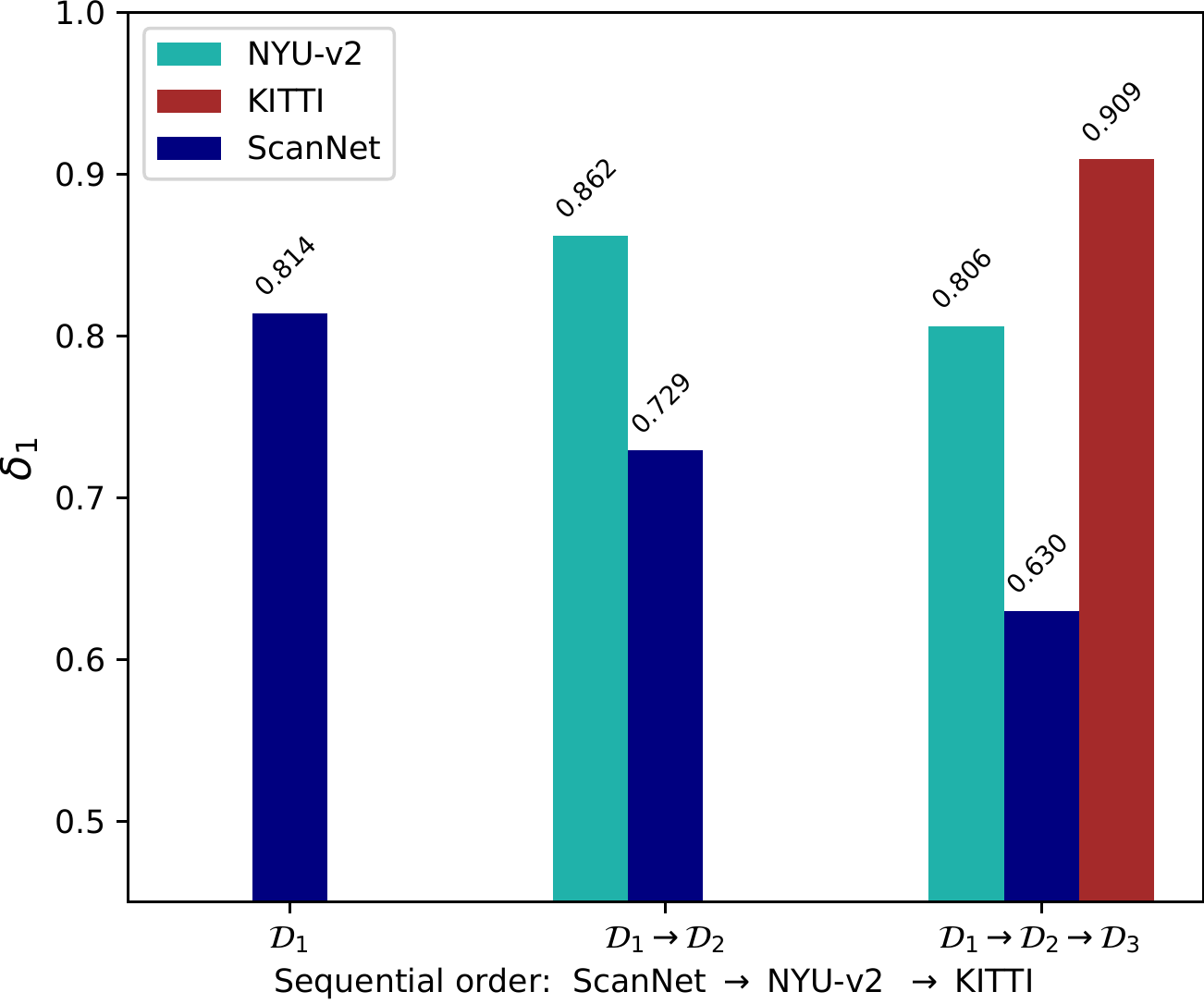}}
\caption{The $\delta_1$ accuracy on three domains for different learning orders.}
\label{fig_vis_results}
\end{figure}

\subsubsection{Results on Three Domains} 
We then perform experiments on all three domains. 
In this case, the first learned domain will suffer from more long-term forgetting.
As both NYU-v2 and ScanNet are composed of indoor scenes, the domain gap between them is small; on the other hand, they have a tremendous domain gap from KITTI. Thus, the domain order performing lifelong learning also affects the final performance. 
For a fair evaluation, we conduct experiments for all possible combinations.
Totally there are six different combinations regards to learning order. We take a standard evaluation protocol for lifelong learning that computes the mean accuracy over multi-domains.

We report the mean $\delta_1$ accuracy in Table~\ref{results_3domains}. As seen, ScanNet $\rightarrow$ NYU-v2 $\rightarrow$ KITTI demonstrates the best accuracy, whereas KITTI $\rightarrow$ NYU-v2 $\rightarrow$ ScanNet shows the worst performance. 
Also, KITTI  $\rightarrow$ ScanNet $\rightarrow$ NYU-v2 leads to penultimate accuracy. It is not surprising since learning on KITTI would cause the model to forget the domain for a longer time.

Fig.~\ref{fig_vis_results} gives a more detailed visualization of performance evolution. It is observed in Fig.~\ref{fig_vis_results} (a) that the performance on KITTI in KITTI $\rightarrow$ NYU-v2 $\rightarrow$ ScanNet degraded significantly compared to other methods. 
The results agree well with our expectations that we should lastly learn KITTI at best.
Overall, Fig.~\ref{fig_vis_results} (c) demonstrates the best trade-off between stability and plasticity as the variance of $\delta_1$ among the three domains is small.

\begin{figure*}[t!]
\centering  
\begin{tabular}
{p{0.09\textwidth}<{\centering}ccc}
(a) Images
&\IncG [height=0.11\textwidth]{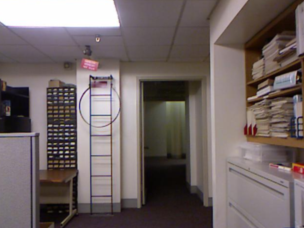} 
&\IncG [height=0.11\textwidth]{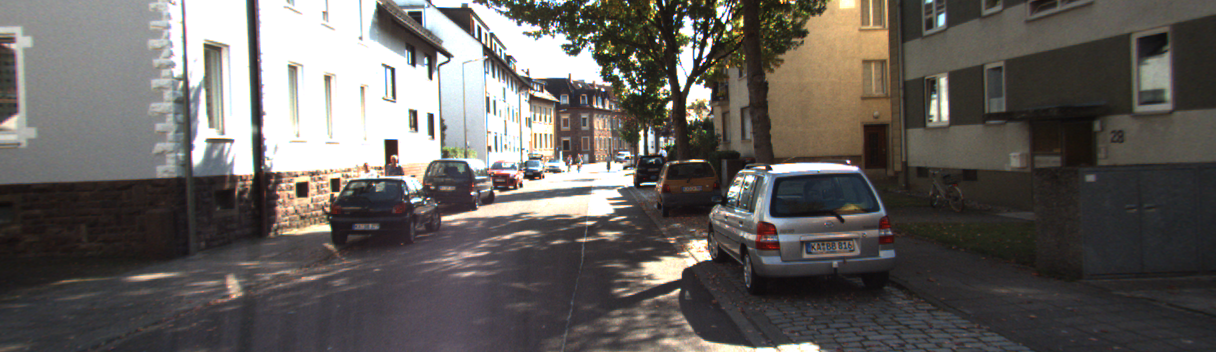} 
&\IncG [height=0.11\textwidth]{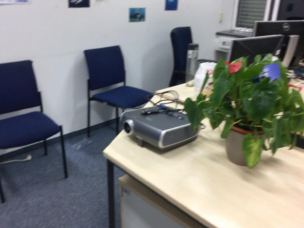} 
\\ 
(b) Ground Truths
&\IncG [height=0.11\textwidth]{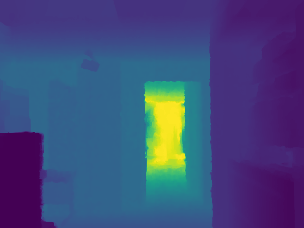} 
&\IncG [height=0.11\textwidth]{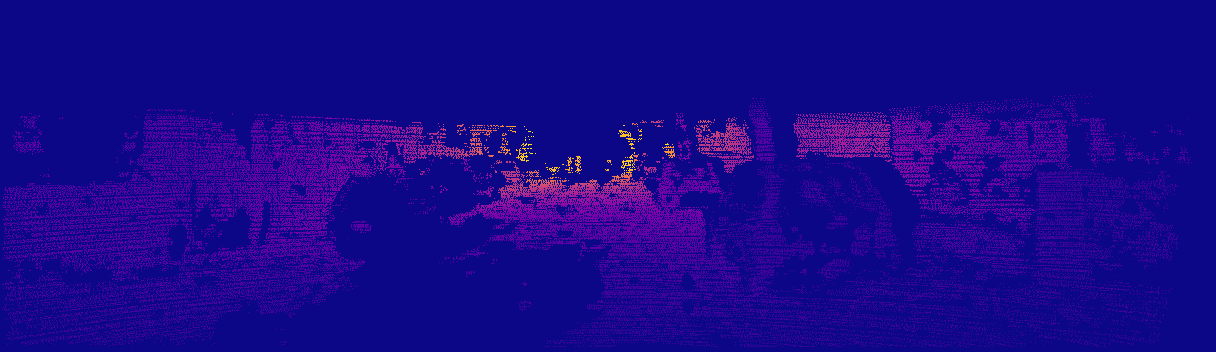} 
&\IncG [height=0.11\textwidth]{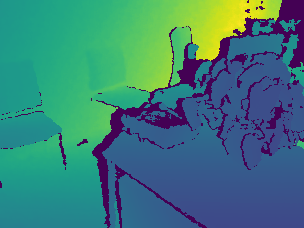} 
\\ 
(c) Estimated depth
&\IncG [height=0.11\textwidth]{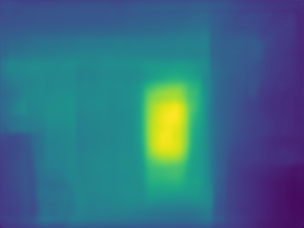} 
&\IncG [height=0.11\textwidth]{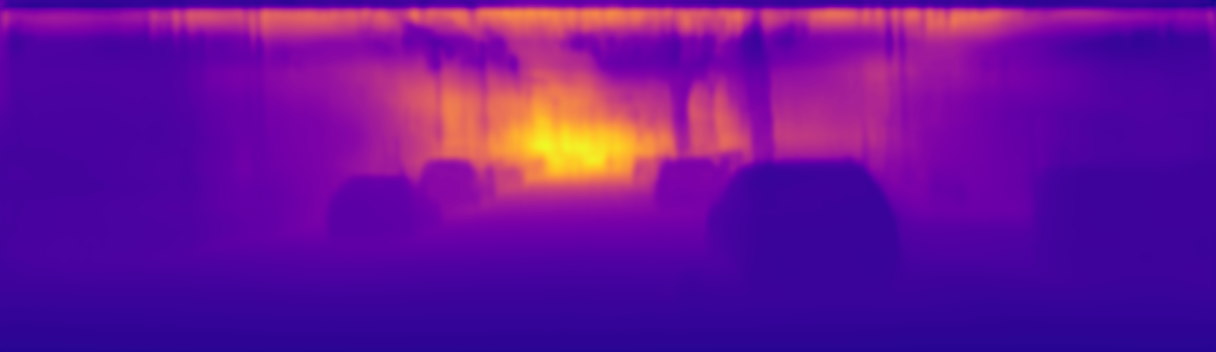} 
&\IncG [height=0.11\textwidth]{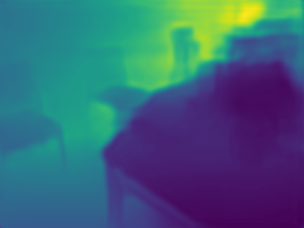} 
\\
(d) Estimated uncertainty
&\IncG [height=0.11\textwidth]{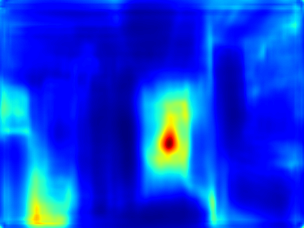} 
&\IncG [height=0.11\textwidth]{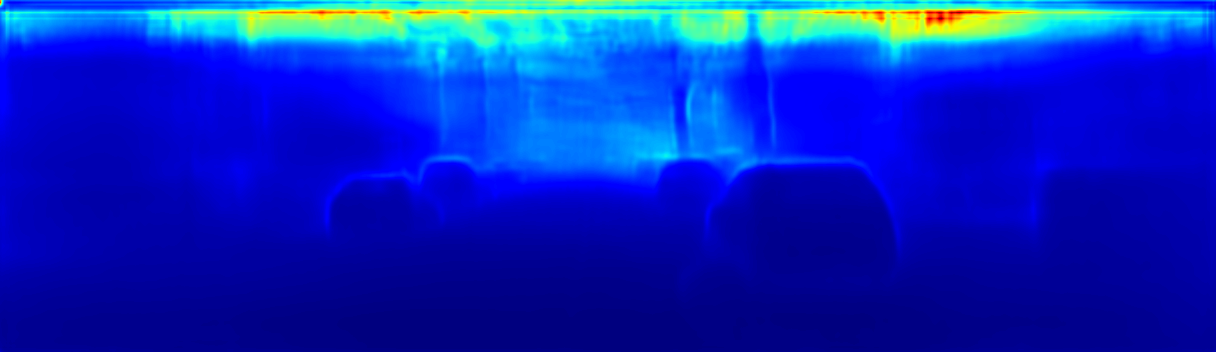} 
&\IncG [height=0.11\textwidth]{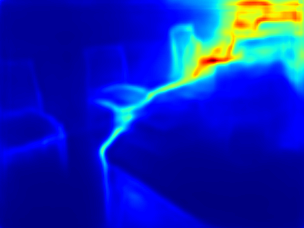} 
\\
&NYU-v2 & KITTI  &ScanNet\\
\end{tabular}
\caption{Qualitative comparison of depth and uncertainty maps predicted by our method in the learning order of NYU-v2 $\rightarrow$ KITTI $\rightarrow$ ScanNet.}
\label{fig-qua-res2}
\end{figure*}

We also provide qualitative results in Fig.~\ref{fig-qua-res2}. As seen, our approach can accurately infer uncertainty maps for the images of the corresponding domain. Generally, high uncertainty is around missing points, far regions, and object boundaries. We will quantify the effects of uncertainty estimation in ablation studies.

\begin{table*}[!t] 
\caption{The $\delta_1$ accuracy of lifelong depth learning on all three domains where domain prior denotes results of manually specify domain-specific predictor for input images. On the contrary, our predictor selection automatically chooses the predictor based on the minimum feature distance.}
\renewcommand\arraystretch{1.2}
\begin{center}
\label{results_online}
\begin{tabular}
{c|ccc|ccc|ccc}
\hline
&\multicolumn{3}{c|}{Domain Prior} &\multicolumn{3}{c|}{Our Predictor Selection} &\multicolumn{3}{c}{Accuracy Drop}  \\ 
Learning order  &NYU-v2 &KITTI & ScanNet & NYU-v2 &KITTI & ScanNet  &  NYU-v2 &KITTI & ScanNet\\ \hline
Ours (NYU-v2 $\rightarrow$ KITTI ) &0.768 &0.910 &- &0.768 &0.910 &- & 0\% &0\% &-\\ 
Ours (KITTI $\rightarrow$ NYU-v2 ) &0.812 &0.813 & - &0.812 &0.813 & - & 0\% &0\% &-\\ 
Ours (NYU-v2 $\rightarrow$ KITTI $\rightarrow$ ScanNet) &0.774 &0.815 &0.751 &0.769 &0.815 &0.748 &0.5\% $\downarrow$ & 0\% &0.3\% $\downarrow$ \\ 
Ours (ScanNet $\rightarrow$ KITTI $\rightarrow$ NYU-v2) &0.837 &0.830 &0.655 &0.805 &0.830 &0.667 & 3.2\% $\downarrow$ & 0\% &0.2\% $\uparrow$\\ 
Ours (KITTI $\rightarrow$ NYU-v2 $\rightarrow$ ScanNet) &0.794 &0.502 &0.784 &0.794 &0.502 &0.764 &0\% & 0\% &2\% $\downarrow$\\ 
Ours (KITTI $\rightarrow$ ScanNet $\rightarrow$ NYU-v2) &0.809 &0.732 &0.645 &0.770 &0.732 &0.651 &3.9\%$\downarrow$ &0\% &0.6\% $\uparrow$\\ 
Ours (ScanNet $\rightarrow$ NYU-v2 $\rightarrow$ KITTI) &0.806 &0.909 &0.630 &0.780 &0.909 &0.639 &2.6\%$\downarrow$ &0\% &0.9\% $\uparrow$\\ 
Ours (NYU-v2 $\rightarrow$ ScanNet $\rightarrow$ KITTI) &0.747 &0.898 &0.693&0.751&0.898&0.672 &0.4\% $\uparrow$ &0\% &2.1\%$\downarrow$\\ 
\hline
\end{tabular}
\end{center}
\end{table*}

\subsection{Results of Online Predictor Selection}
\begin{figure}[t]
\centering
\subfigure[\footnotesize{NYU-v2 $\rightarrow$  KITTI $\rightarrow$ ScanNet }]
{\includegraphics[width=0.22\textwidth]{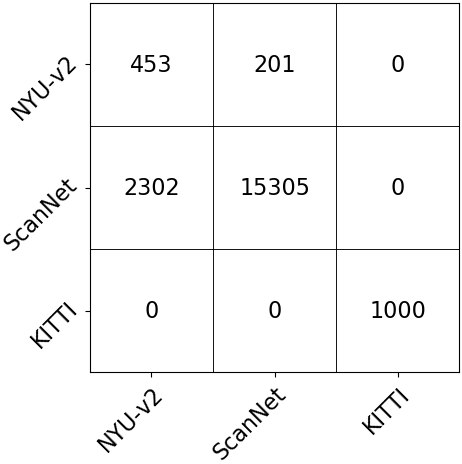}}
\subfigure[\footnotesize{KITTI $\rightarrow$ NYU-v2 $\rightarrow$ ScanNet}]
{\includegraphics[width=0.22\textwidth]{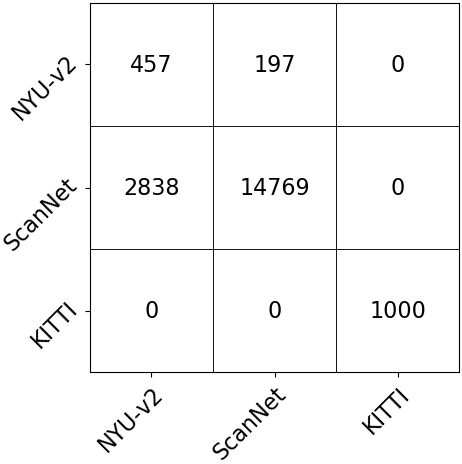}}
\vspace{-2mm}
\caption{Results of the number of categorized images on test sets of the three domains.}
\label{fig_vis_confusion}
\end{figure}

Online predictor selection is a relatively practical requirement. Since there are multiple depth predictors, given an input image, the model must automatically identify its domain and select the correct branch to infer a depth map. Therefore, we conduct experiments to validate the effectiveness of the proposed predictor selection method for the trained framework varying the learning order. The results are given in Table~\ref{results_online} in which Domain Prior denotes the results of pre-specifying the corresponding predictor for input images. It provides an upper bound to our predictor selection method. As seen, for results of lifelong learning on two domains, \ie, NYU-v2 $\rightarrow$ KITTI and KITTI $\rightarrow$ NYU-v2, our predictor selection method demonstrates a 100\% success rate. For results on three domains, our method still attained 100\% success rate for KITTI while yielding a slight accuracy drop (within 4\%) for NYU-v2 or ScanNet. We consider the miss between NYU-v2 and ScanNet reasonable as they contain some similar indoor images. It can be better observed in Fig.~\ref{fig_vis_confusion}, which shows the number of categorized images on test sets of the three domains. Fig.~\ref{fig_vis_confusion} (a) and (b) show results of Ours (NYU-v2 $\rightarrow$ KITTI $\rightarrow$ ScanNet) and Ours (KITTI $\rightarrow$ NYU-v2 $\rightarrow$ ScanNet), respectively. It is seen that our predictor selection method could identify data from KITTI without misclassification. Although there are some misclassified images between NYU-v2 and ScanNet, as we discussed above, the depth maps can still be accurately inferred due to the small domain gap. Therefore, the accuracy drop is slight and acceptable. Besides, as shown in Table~\ref{results_online}, the accuracy could be improved sometimes.

We also evaluate the computational efficiency of our proposed method, running on a computer with Intel(R) Xeon(R) Gold 6230 CPU and a GeForce RTX 2080 Ti GPU card.
We calculate the GPU time by running the model for an input image 10000 times and calculate the mean time. 
Table~\ref{results_efficiency} shows the results for the three domains. As seen, our predictor selection module spends only 2.7 ms more time for NYU-v2 and ScanNet, and 0.5 ms for KITTI, respectively, demonstrating superior efficiency.

\begin{table}[t]
\caption{Results of computational efficiency.}
\renewcommand\arraystretch{1.2}
\begin{center}
\label{results_efficiency}
\begin{tabular}
{p{0.08\textwidth}<{\centering}|p{0.08\textwidth}<{\centering}p{0.1\textwidth}<{\centering}p{0.1\textwidth}<{\centering}}
\hline
\multirow{3}{*}{Datasets} & \multirow{3}{*}{Resolution} & GPU [ms] w/o predictor selection & GPU [ms] w predictor selection  \\ \hline
NYU-v2 & 304 $\times$ 228 & 8.2 & 10.9\\
ScanNet &304 $\times$ 228 & 8.2 &10.9\\
KITTI &1216 $\times$ 352 &28.0 &28.5\\
\hline
\end{tabular}
\end{center}
\end{table}

\subsection{Ablation Study}
We perform several ablation studies to analyze our method better. For simplicity, we conduct experiments on two domains. We take ours (NYU-v2 $\rightarrow$ KITTI) as the base method and remove some critical operations, including data replay, uncertainty consistency, and scale-invariant operation, for comparison. The results are given in Table~\ref{results_as}.

\textbf{Without uncertainty estimation}: the uncertainty is used in the uncertainty-aware loss Eq.\eqref{un_loss} and consistency loss Eq.\eqref{kd_loss}.
We remove the uncertainty estimation module to evaluate the performance. As a result, we observe performance degradation both on NYU-v2 and KITTI. 

\textbf{Without data replay}: data replay is used to enhance the stability of the model. The results without replay demonstrate 0.3\% and 2\% accuracy drop on KITTI and NYU-v2, respectively.
It indicates that replay is more important in improving stability.

\textbf{Without uncertainty consistency}: The uncertainty consistency is applied along with depth consistency in the original method to prevent forgetting. As shown,
 without uncertainty consistency, the performance further degrades mainly for the old domain, even though we observe a slight improvement for the new domain.
 
 \textbf{With a different backbone network}:
We replace the ResNet-34 based encoder with MobileNet-v2 \cite{sandler2018mobilenetv2}. It gives us a more lightweight network with only 1.99 M parameters.
The $\delta_1$ accuracy is 0.733 and 0.901 for NYU-v2 and KITTI, respectively, and the mean accuracy reaches 0.817, which still outperforms other baseline methods built on large networks.

\begin{table}[t]
\caption{Results of ablation studies.}
\renewcommand\arraystretch{1.2}
\begin{center}
\label{results_as}
\begin{tabular}
{c|ccc}
\hline
Method & NYU-v2 & KITTI &Average\\ \hline
Ours (NYU-v2 $\rightarrow$ KITTI) &0.768 &0.910 &0.839\\
w/o uncertainty estimation &0.740 & 0.857 &0.799\\
w/o $\ell_{replay}$ &0.749 &0.907 &0.828\\
w/o $\ell_{replay}$ and uncertainty consistency &0.700 &0.913 &0.807\\
\hline
\end{tabular}
\end{center}
\end{table}

\subsection{Summary}

\begin{itemize}
    \item The proposed multi-head lifelong depth learning framework, \ie, {\it Lifelong-MonoDepth}, can estimate depth maps with the absolute scale from multi-domains even though there exist significant domain gaps. 
    \item {\it Lifelong-MonoDepth} attains a good balance between stability and plasticity on real-world datasets. It generally outperforms baseline methods by around 8\% $\sim$ 15\%.
     \item {\it Lifelong-MonoDepth} can automatically identify the domain-specific predictor during inference, showing satisfactory accuracy and efficiency.
     \item The learning order of domains has an essential effect on lifelong depth learning. For example, learning in NYU-v2 $\rightarrow$ ScanNet $\rightarrow$ KITTI substantially outperforms KITTI $\rightarrow$ NYU-v2 $\rightarrow$ ScanNet in average accuracy over multi-domains. 
     Generally, learning in an indoor $\rightarrow$ outdoor order contributes to better performance.
     In practice, the learning order should be decided according to the specific applications.
\end{itemize}

\section{Conclusion}
\label{conclusion}

We present a novel lifelong learning framework for multi-domain metric depth estimation, namely {\it Lifelong-MonoDepth}. We argue that the major challenges are i) large domain gaps and ii) depth scale imbalance, which cause catastrophic forgetting in lifelong learning. We then propose an efficient multi-head network composed of a domain-shared encoder and domain-specific predictors. Such multi-head predictors enable estimate depth maps with different scales and mitigate domain shift.
To alleviate catastrophic forgetting, we propose a novel strategy that applies both depth and uncertainty consistency to avoid knowledge forgetting and uses replay regularization to improve stability further.

We conduct extensive numerical studies to demonstrate the effectiveness of our method. We 
show that our approach outperforms all baseline methods by a good margin. We also provide the effects of varying the learning order of multiple domains.
During inference, we propose to calculate the distance between an image and each domain; then, the minimum distance corresponds to the domain-specific predictor to infer a depth map. 

For the first time, we are able to enable scale-aware depth prediction across multi-domains with significant domain gaps in lifelong learning. Potential applications of our method include visual navigation, obstacle avoidance, 3D perception. We hope our method can inspire more future explorations on lifelong depth learning.

\IEEEpeerreviewmaketitle

\bibliographystyle{IEEEtranS}
\bibliography{egbib}

\end{document}